%% file: main.tex
\documentclass[runningheads]{llncs}
\usepackage{graphicx}
\usepackage{tikz}
\usepackage{comment}
\usepackage{amsmath,amssymb} 
\usepackage{color}
\input{preamble}

\usepackage[accsupp]{axessibility} 
\usepackage{subcaption}

    \makeatletter
\def\@fnsymbol#1{\ensuremath{\ifcase#1\or \dagger \or * \or \ddagger\or
   \mathsection\or \mathparagraph\or \|\or **\or \dagger\dagger
   \or \ddagger\ddagger \else\@ctrerr\fi}}
    \makeatother

\begin{document}
\pagestyle{headings}
\mainmatter
\def\ECCVSubNumber{1243}

\title{Long-Tailed Class Incremental Learning} %

\titlerunning{Long-Tailed Class Incremental Learning}

\author{Xialei Liu\inst{1,*,} \thanks{Corresponding author (xialei@nankai.edu.cn)}
\and
Yu-Song Hu\inst{1,} \thanks{The first two authors contribute equally.}
\and
Xu-Sheng Cao\inst{1}
\and
Andrew D. Bagdanov \inst{2}
\and
Ke Li\inst{3}
\and
Ming-Ming Cheng\inst{1}
}
\authorrunning{X. Liu et al.}
\institute{TMCC, CS, Nankai University, China\inst{1}; MICC, University of Florence,
Florence, Italy\inst{2} Tencent Youtu Lab\inst{3}}
\maketitle

\begin{abstract}

In class incremental learning (CIL) a model must learn new classes in a sequential manner without forgetting old ones.
However, conventional CIL methods consider a balanced distribution 
for each new task, which ignores the prevalence of long-tailed distributions 
in the real world. 
In this work we propose two long-tailed CIL scenarios,
which we term \emph{ordered} and \emph{shuffled} LT-CIL. 
\emph{Ordered} LT-CIL considers the scenario where we learn from head classes collected with more samples than tail classes which have few. 
\emph{Shuffled} LT-CIL, on the other hand, assumes a completely random long-tailed distribution for each task. We systematically evaluate existing methods in both LT-CIL scenarios and demonstrate very different behaviors compared to conventional CIL scenarios.
Additionally, we propose a two-stage learning baseline with a learnable weight scaling layer for reducing the bias caused by long-tailed distribution in LT-CIL and which in turn also improves the performance of conventional CIL due to the limited exemplars. 
Our results demonstrate the superior performance 
(up to 6.44 points in average incremental accuracy) of our approach on CIFAR-100 and ImageNet-Subset. 
The code is available at \footnotesize\url{https://github.com/xialeiliu/Long-Tailed-CIL}.
\end{abstract}

\section{Introduction}\label{sec:intro}
\input{intro}

\section{Related Work}\label{sec:rel}
\input{rel}

\section{A Two-stage Approach to LT-CIL}
\label{sec:method}
\input{method}

\section{Experimental Results}\label{sec:exp}
\input{experiments}

\section{Conclusions}\label{sec:con}
\input{conclusion}

\paragraph{Acknowledgments}
This work is funded by National Key Research and the Development Program of China (NO. 2018AAA0100400), NSFC (NO. 61922046), NSFC (NO. 62206135) the S\&T innovation project from the Chinese Ministry of Education, and by the European Commission under the Horizon 2020 Programme, grant number 951911 -- AI4Media.

\clearpage
\bibliographystyle{splncs04}
\bibliography{egbib}
\newpage
\appendix
\section{More results on ImageNet-Subset}
\begin{figure*}[t]
\begin{minipage}[b]{0.32\linewidth}
\centering
\includegraphics[width=\textwidth]{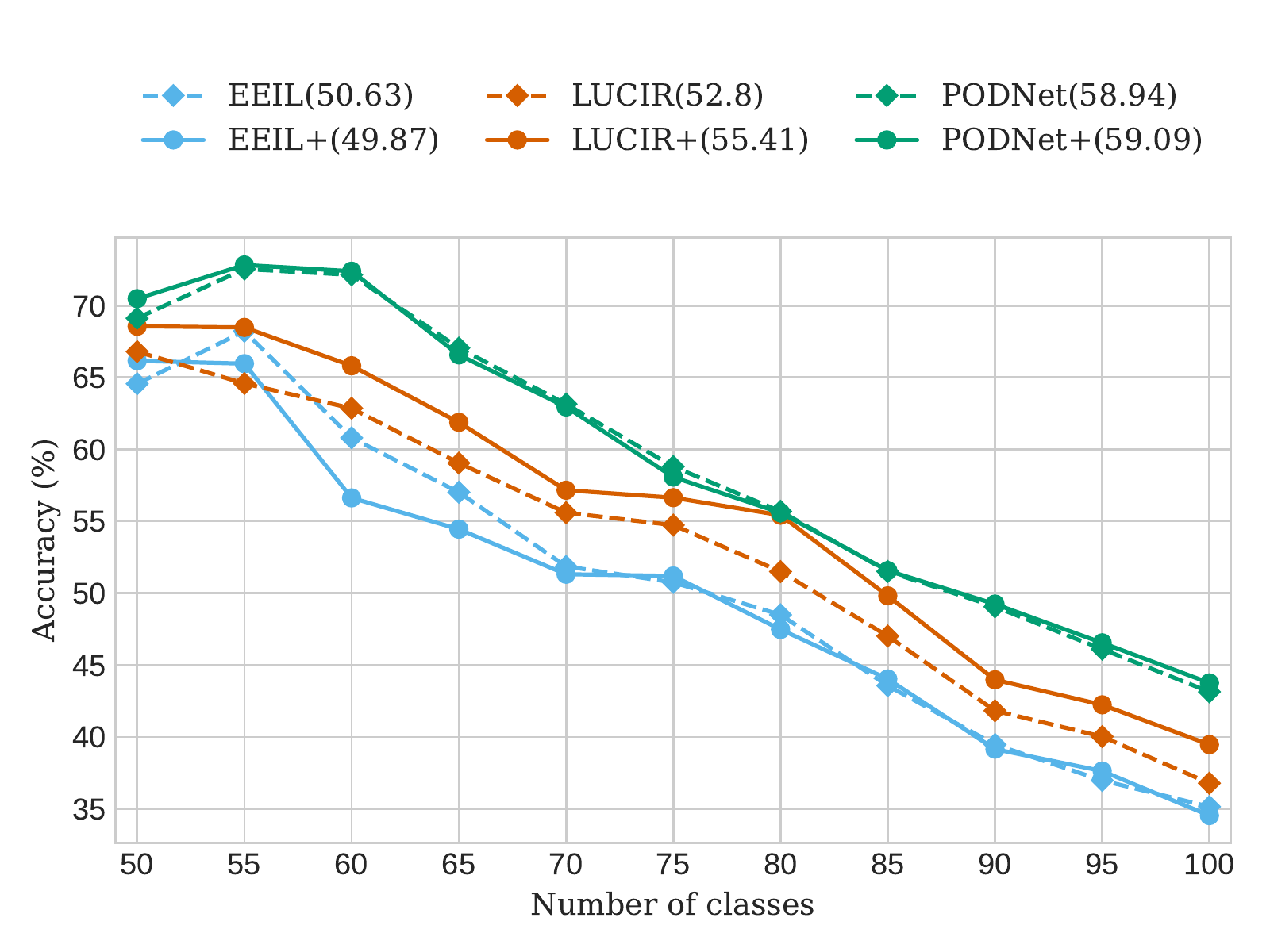}
\subcaption{$\rho$ = 0.01 ordered LT}
\end{minipage}
\begin{minipage}[b]{0.32\linewidth}
\centering
\includegraphics[width=\textwidth]{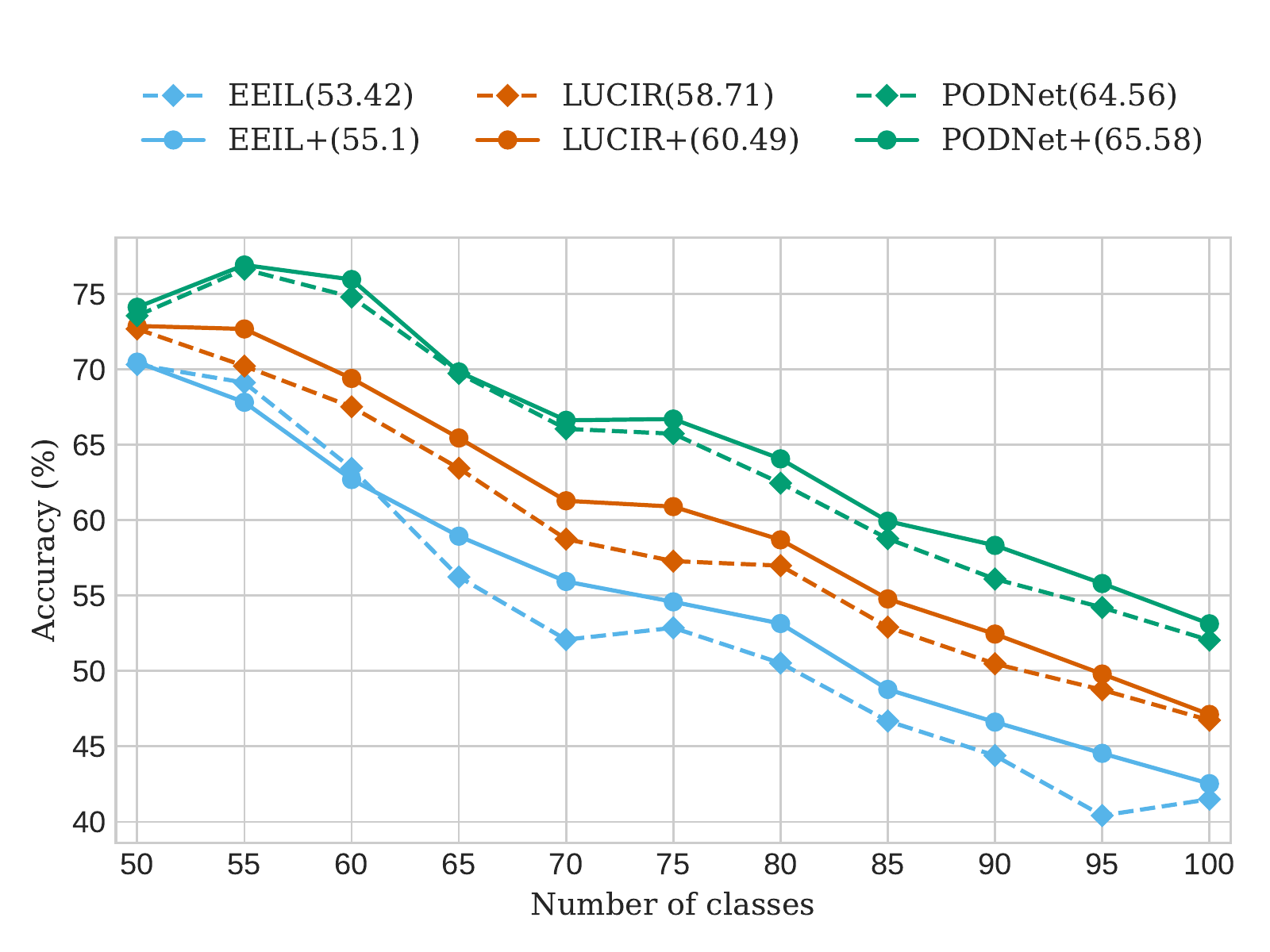}
\subcaption{$\rho$ = 0.05 ordered LT}
\end{minipage}
\begin{minipage}[b]{0.32\linewidth}
\centering
\includegraphics[width=\textwidth]{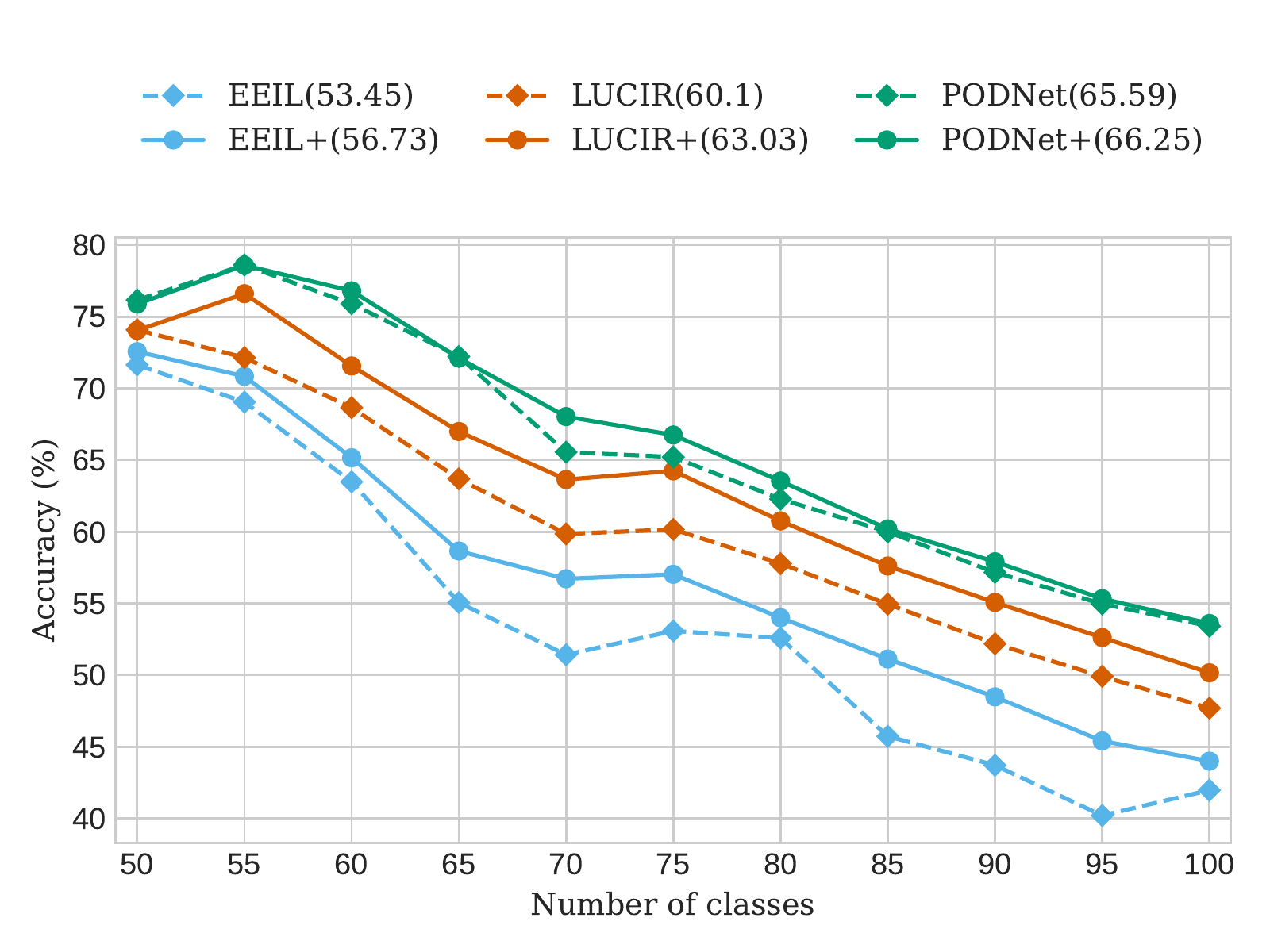}
\subcaption{$\rho$ = 0.1 ordered LT}
\end{minipage}
\begin{minipage}[b]{0.32\linewidth}
\centering
\includegraphics[width=\textwidth]{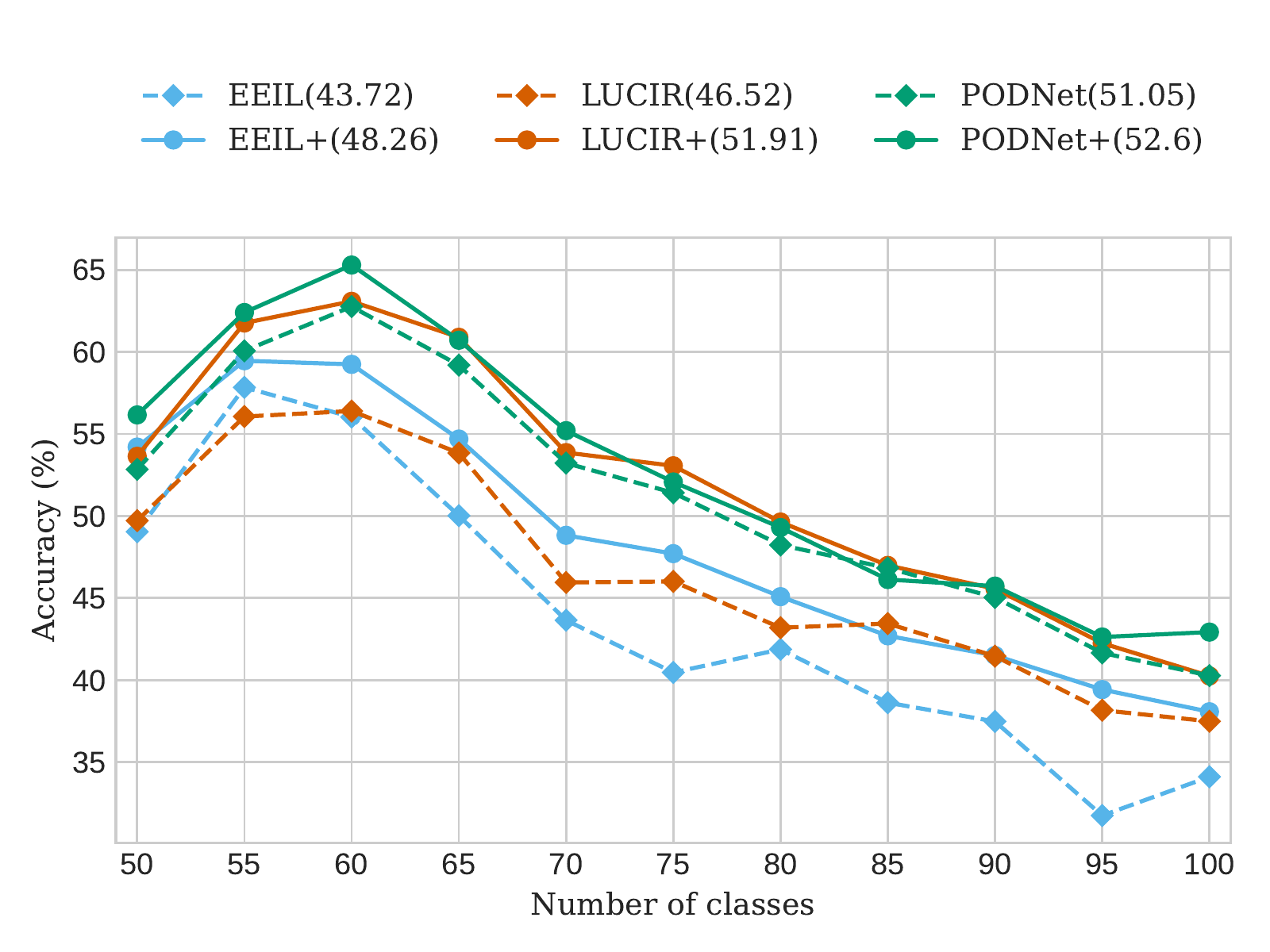}
\subcaption{$\rho$ = 0.01 shuffled LT}
\end{minipage}
\begin{minipage}[b]{0.32\linewidth}
\centering
\includegraphics[width=\textwidth]{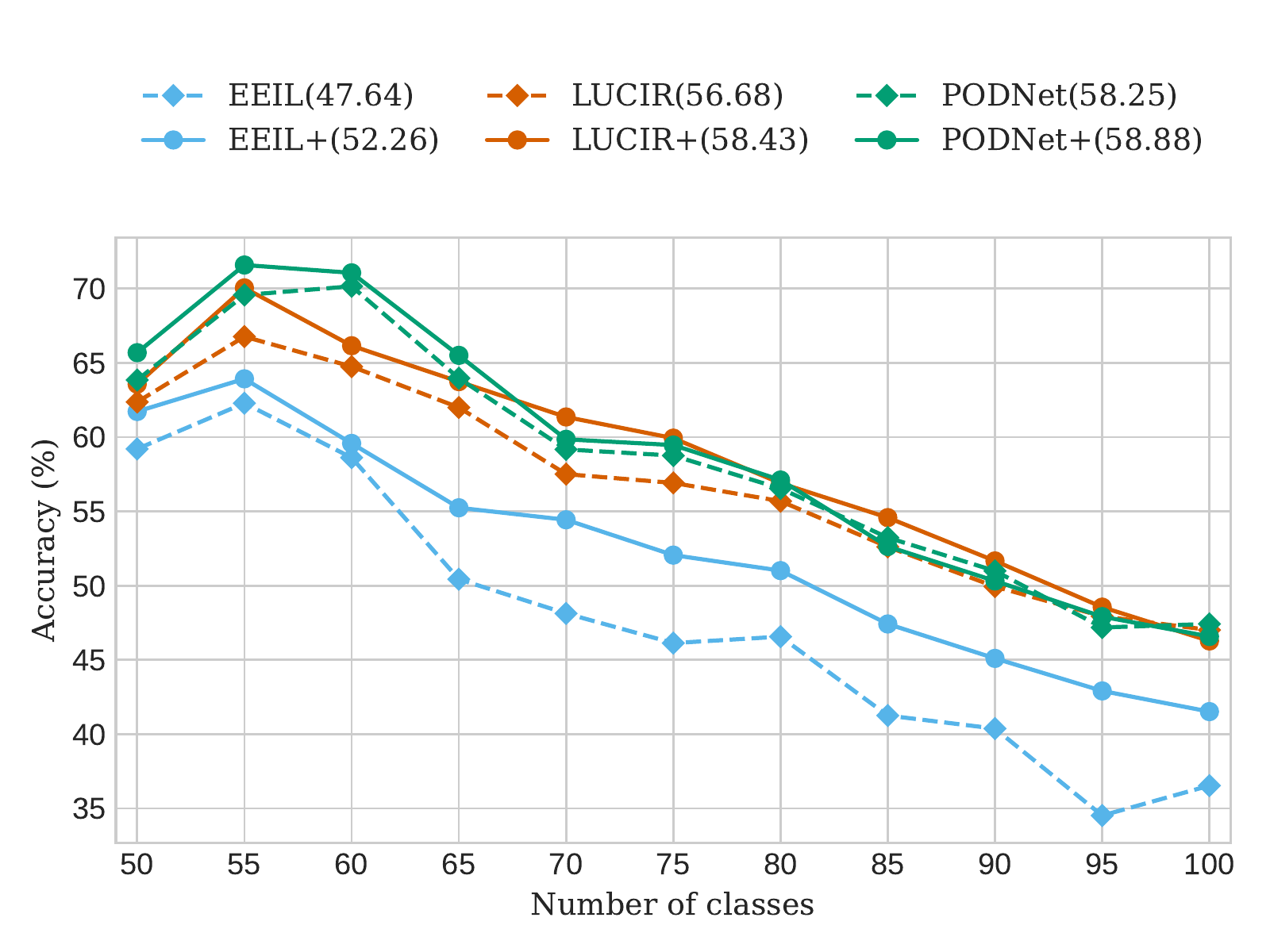}
\subcaption{$\rho$ = 0.05 shuffled LT}
\end{minipage}
\begin{minipage}[b]{0.32\linewidth}
\centering
\includegraphics[width=\textwidth]{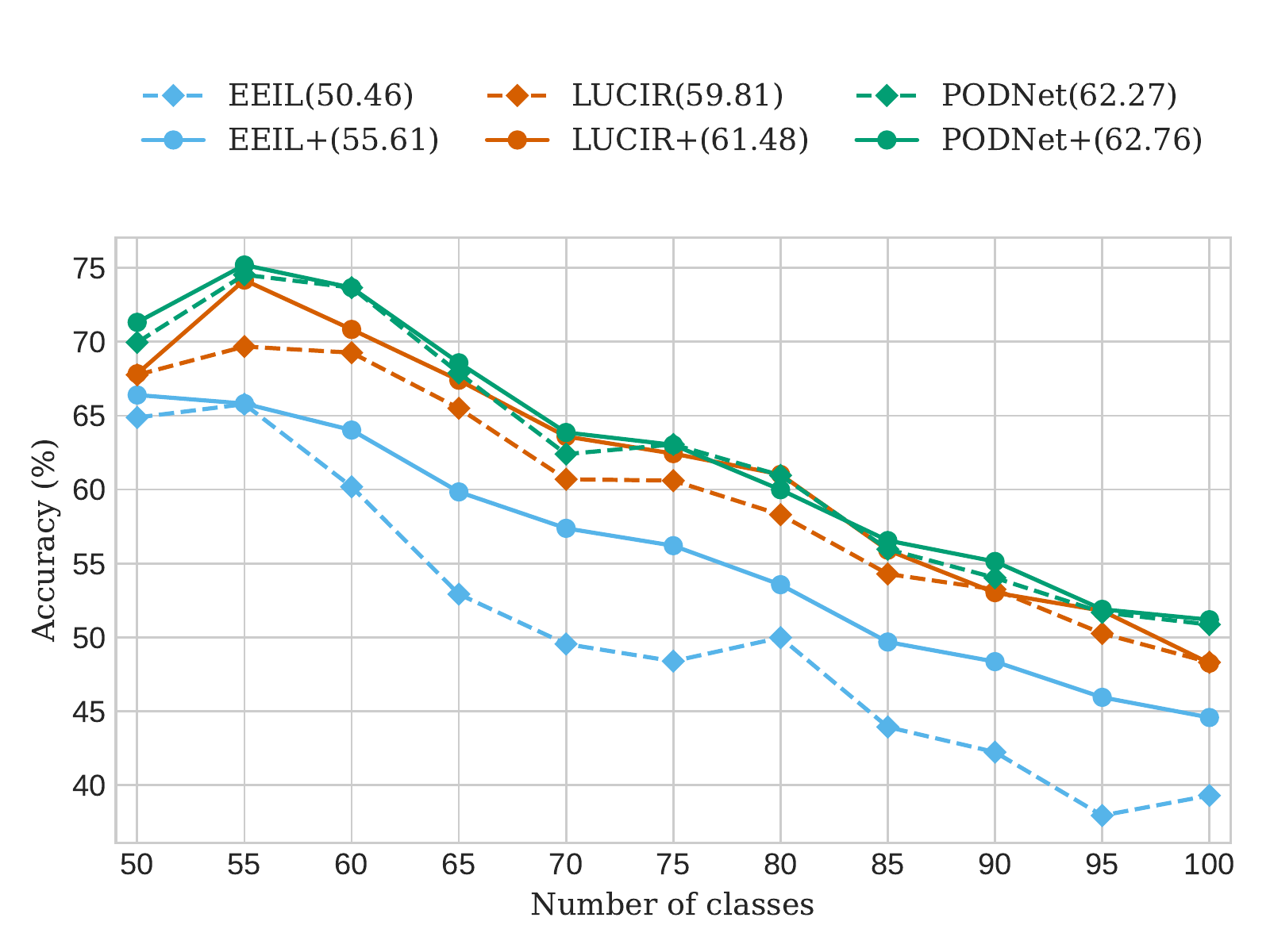}
\subcaption{$\rho$ = 0.1 shuffled LT}
\end{minipage}
\caption{Average accuracy on different imbalance ratios on Imagenet-Subset, the top row is on ordered LT-CIL and the bottom row is on shuffled LT-CIL. Methods with $+$ sign indicate our two-stage method applied to the corresponding baseline.
}
\label{fig:imb_ratio_imagenet}
\end{figure*}

\paragraph{Different imbalance ratios on ImageNet-Subset}
 In this section, we analyze three different imbalance ratios $\rho=0.01$, $\rho=0.05$ and $\rho=0.1$ on ImageNet-Subset, the smaller the ratio the more skewed the distribution. Compared to CIFAR100, ImageNet-Subset contains more samples, which results in a more skewed distribution on different continual training steps. We report three baselines, i.e. EEIL, LUCIR and PODNET, and our two-stage approach applied to them, denoted as EEIL+, LUCIR+ and PODNET+. As we can see from Figure~\ref{fig:imb_ratio_imagenet} (a-c), with more samples, for the ordered LT-CIL scenario, PODNET surpasses other approaches consistently with a large margin, obtaining the best performance in all scenarios. We consider that PODNET can learn much more information when the data is sufficient. Overall our two-stage method can consistently boost accuracy for most methods, especially for LUCIR+ with a significant gain. For shuffled LT-CIL scenario from Figure~\ref{fig:imb_ratio_imagenet} (d-f), PODNET+ and LUCIR+ are very competitive in all three imbalance ratio $\rho$. The proposed two-stage method further improves the performance, especially for EEIL and LUCIR.
Interestingly, we can see that compared to conventional settings, long-tailed scenarios with a large imbalance ratio can achieve competitive performance with less samples, which may due to the imbalance effect of training data.
\begin{table}[tb]\Large
\setlength{\tabcolsep}{1pt}{
\resizebox{0.8\textwidth}{!}{
    \begin{tabular*}{395pt}{l|cc|cc|cc}
    \toprule
                 & EEIL  & EEIL+ & LUCIR & LUCIR+         & PODNet & PODNet+        \\ \midrule
    Conventional & 44.14 & 55.92 & 53.60 & 56.96          & 60.25  & \textbf{64.75} \\
    Ordered      & 47.18 & 46.43 & 47.05 & 49.77 & \textbf{58.30}  & 57.76          \\
    Shuffled     & 40.75  & 43.95 & 40.52 & 48.35 & 48.93 & \textbf{51.52}         \\ \bottomrule
    \end{tabular*}}
}
\caption{The average accuracy on long sequence of 25 steps for three different scenarios on Imagenet-Subset. Methods with $+$ sign indicate our two-stage method applied to the corresponding baseline.}
\label{tab:long_imagenet}
\end{table}
\paragraph{Long sequence on ImageNet-Subset}
    We evaluate on long sequence of 25 steps for all three scenarios with three state-of-the-art methods on ImageNet-Subset, and collect the results in Table~\ref{tab:long_imagenet}. As we can see, our method also  improves over different  baselines in this more challenging setting like on CIFAR100 except for PODNET on ordered LT-CIL scenario. Further more, we can see that for 25-step scenario, two-stage methods can get much larger gain than in 5-step and 10-step scenarios in most cases. It shows that the two-stage methods are more robust for longer sequences.

\end{document}

%% file: preamble.tex
\usepackage{amssymb}
\usepackage{bm}
\usepackage{multirow}
\usepackage[ruled,vlined]{algorithm2e}
\usepackage{floatrow}

\usepackage{booktabs}
\usepackage{xspace}

\usepackage[pagebackref,breaklinks,colorlinks,bookmarks=false]{hyperref}
\usepackage[capitalise]{cleveref}
\usepackage{url}

\usepackage{graphicx}
\usepackage{lipsum}
\usepackage{microtype}
\usepackage{nicefrac}
\usepackage{verbatim}
\usepackage{bbding}

\makeatletter
\DeclareRobustCommand\onedot{\futurelet\@let@token\@onedot}
\def\@onedot{\ifx\@let@token.\else.\null\fi\xspace}

\renewcommand{\paragraph}{%
	\@startsection{paragraph}{4}{\z@}
	{0.1em \@plus 0.5ex \@minus 0.2ex}{-1em}%
	{\normalsize\bf}%
}
\makeatother

%% file: intro.tex
Deep neural networks have achieved spectacular success in many computer vision tasks. In general, most tasks assume a static world in which all data is available for training in a single learning session. The world is ever-changing, however, and future intelligent systems will have to master new tasks and adapt to new environments without forgetting previously acquired knowledge. Incremental learning, also known as continual or lifelong learning, is the paradigm of continually learning a sequence of tasks as new data becomes available~\cite{belouadah2020comprehensive,delange2021continual,masana2020class,parisi2019continual}. The biggest challenge in incremental learning is avoiding \emph{catastrophic forgetting}~\cite{mccloskey1989catastrophic} when learning with only current task data and possibly a small memory of data from previous tasks.

Class incremental learning (CIL) considers a scenario in which no task boundary is provided during inference, which is significantly more challenging than task incremental learning with a known task boundary~\cite{van2019three}. 

Although conventional CIL has seen significant progress, it assumes that the data is sampled from a balanced distribution. However, data sampled from the real world often follows a long-tailed distribution~\cite{huang2019deep,liu2019large,wang2017learning,zhong2019unequal} in which some classes have many more samples than others. Learning from long-tailed distributions has been approach by re-sampling~\cite{han2005borderline,japkowicz2002class} or re-weighting~\cite{lin2017focal,shu2019meta,zhang2017range} head classes and tail classes to learn a balanced classifier. Recently, transfer learning~\cite{liu2019large,yin2019feature} between head and tail classes, two-stage learning~\cite{kang2019decoupling,zhong2021improving} to decouple representation and classifier learning, and ensemble learning~\cite{xiang2020learning,wang2020long} of different experts have achieved superior performance. However, all these works consider a static world in which all data is immediately available for training. It is not straightforward to extend these methods with new classes without suffering catastrophic forgetting. 
\begin{figure}[t]
\resizebox{0.8\columnwidth}{!}{%
\includegraphics[width=\textwidth]{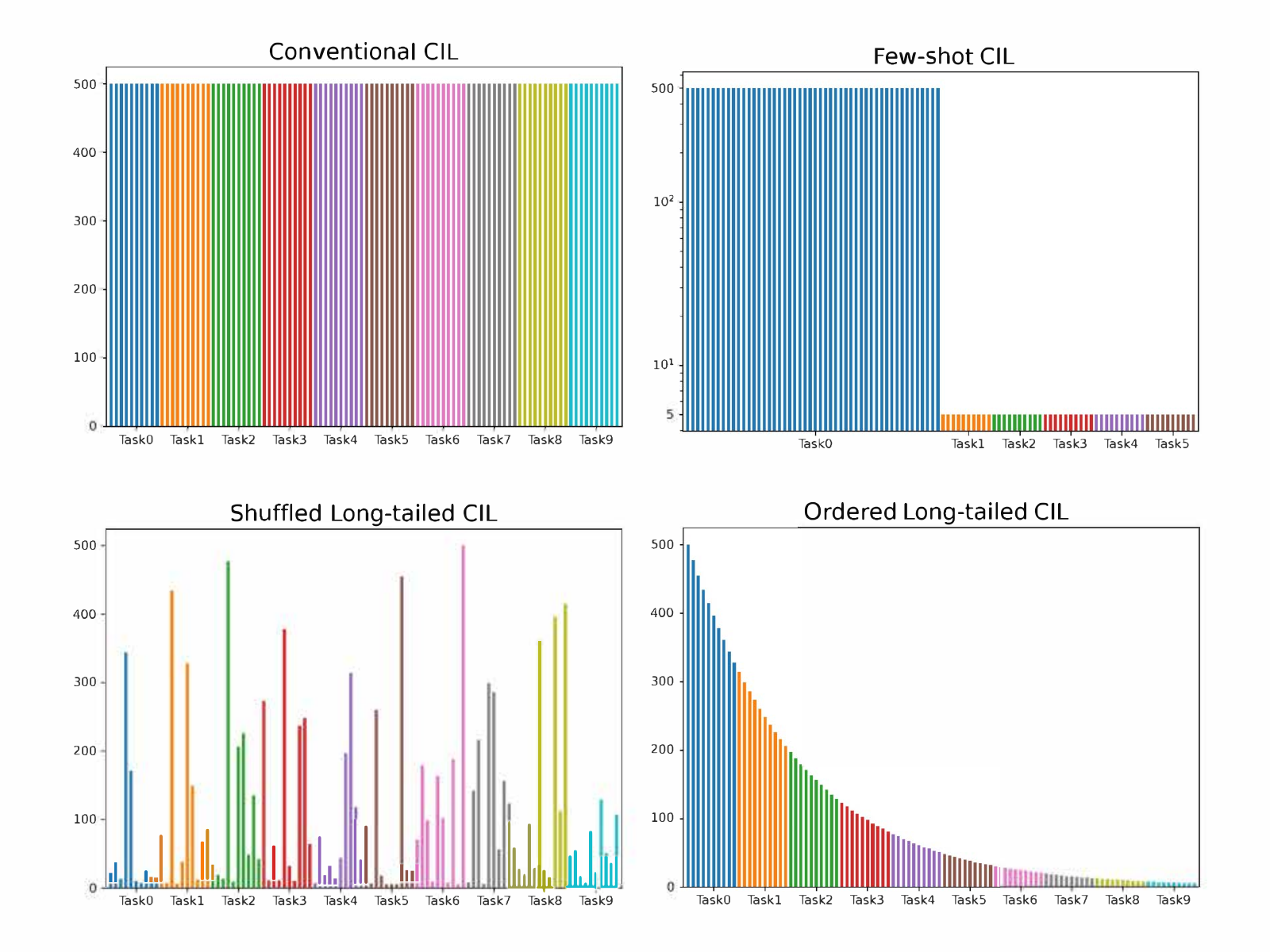}
}
\caption{An illustration of our proposed long-tailed CIL (LT-CIL) scenarios compared to conventional CIL~\cite{masana2020class} with balanced distribution and few-shot CIL~\cite{tao2020few}.}

\label{fig:longtail}
\end{figure}

Due to the long-tailed and incremental nature of real-world learning problems, it is crucial to investigate the class incremental learning in the more realistic scenario with long-tailed distribution for different tasks. In this work, we propose two new scenarios based on long-tailed distributions: \emph{Ordered} Long-tailed CIL (LT-CIL) and \emph{Shuffled} Long-tailed CIL. As shown in Fig.~\ref{fig:longtail}, \emph{Ordered} LT-CIL considers a scenario in which all classes are ordered according to the number of samples per class and then they are divided into different tasks. In contrast, \emph{Shuffled} LT-CIL assumes that classes appearing in different tasks are random and each task may have varying degrees of imbalance to their distributions. Compared to current common CIL and few-shot CIL scenarios, LT-CIL considers more natural data distributions from the real-world.

We compare existing state-of-the-art CIL algorithms on these new LT-CIL scenarios, and our results show that they all perform much worse under long-tailed distributions. They are also less robust to different datasets and less consistent with increasing number of exemplars compared to conventional CIL. Therefore, we propose a two-stage strategy with a learnable weight scaling layer for LT-CIL to boost the performance of existing methods. 
As an extra bonus, we find that the two-stage strategy can also help on the conventional CIL scenario, where a limited memory for previous data and the large amount of current data can cause unbalanced data distribution as well. Importantly, our two-stage approach can be integrated into any CIL method.

The main contributions of this paper are:
\begin{itemize}
    \item we propose two new CIL scenarios (Ordered and Shuffled LT-CIL) that consider long-tailed class distributions more common in the real world;
    
    \item we evaluate conventional CIL algorithms comprehensively and report several findings in these two new long-tailed scenarios; 
    
    \item we design a two-stage training strategy with a learnable weight scaling layer for LT-CIL scenarios which is complementary to existing CIL methods and show that it improves conventional CIL and both LT-CIL scenarios on CIFAR-100 and ImageNet. 
    
\end{itemize}

%% file: rel.tex
Here we review recent work from the literature on class incremental and long-tailed learning most relevant to our proposed approach.

\subsection{Class incremental learning}

Class incremental learning (CIL) is one of the primary scenarios for continual learning~\cite{van2019three}. There are three main approaches to tackling this problem: regularization-based methods, parameter-isolation methods, and replay-based methods. Elastic Weight Consolidation (EWC) is a popular regularization-based method which identifies which parameters are more important for previous tasks and updating these less during learning of new tasks~\cite{kirkpatrick2017overcoming}. R-EWC~\cite{liu2018rotate}, Synaptic Intelligence (SI)~\cite{zenke2017continual}, and Memory Aware Synapses (MAS)~\cite{aljundi2018memory} adopt the same strategy but with different techniques to identify important weights. 
Learning without Forgetting (LwF) is a widely-used baseline that uses knowledge distillation technique to constrain the output probabilities of new tasks~\cite{li2018learning}.

Parameter-isolation methods increase model plasticity by adding more neurons, modules~\cite{rajasegaran2019random,schwarz2018progress}, branches ~\cite{liu2021adaptive} or masks~\cite{mallya2018packnet,mallya2018piggyback,serra2018overcoming}. Dynamically Expandable Networks (DEN)~\cite{lee2017lifelong} performs selective retraining and dynamically expands network capacity, while Dark Experience Replay (DER)~\cite{yan2021dynamically} dynamically expands the representation by freezing the previously-learned representation and augmenting it with additional feature dimensions from a new learnable feature extractor.

Replay-based methods are very effective and recall knowledge from previous tasks by maintaining a small memory of samples~\cite{ahn2021ss,belouadah2019il2m,castro2018end,hou2019learning,rebuffi2017icarl,wu2019large}, representations~\cite{hayes2020remind}, or synthetic data~\cite{shin2017continual}. Incremental Classifier and Representation Learning (iCaRL) stores a fixed budget of exemplars to train and construct class means for classification~\cite{rebuffi2017icarl}.  Pooled Output Distillation (PODNET) applies various pooling operations to intermediate features to distill knowledge from past tasks~\cite{douillard2020podnet}.

These conventional CIL approaches all implicitly assume a balanced label distribution for each task. Recently, Kim et al.~\cite{kim2020imbalanced} proposed a multi-label classification problem with long-tailed distribution. Abdelsalam et al.~\cite{abdelsalam2021iirc} proposed another realistic CIL setting in which each class can have two granularity levels: each sample could have a high-level (coarse) label and a low-level (fine) label, but only one label is available for each task. In contrast to these works, we are interested in more realistic scenarios for CIL with long-tailed class distributions. We provide a comprehensive experimental evaluation of state-of-the-art CIL methods on such settings. Additionally, we propose a two-stage framework with a learnable weight scaling layer to further reduce the bias problem caused long-tailed distribution.

\subsection{Long-tailed learning}
The long-tailed learning problem has been comprehensively studied given the prevalence of the data imbalance problem in the real world~\cite{huang2019deep,wang2017learning,zhong2019unequal,liu2019large}. Most previous works address this problem by re-sampling~\cite{han2005borderline,japkowicz2002class}, re-weighting~\cite{lin2017focal,shu2019meta,zhang2017range} or transfer learning~\cite{liu2019large,yin2019feature}.  Re-sampling methods can over-sample the tail classes or under-sample the head classes. Re-weighting methods assign different weights to different classes or instances.  Transfer learning aims to fuse knowledge between head and tail classes. Data augmentation is another way to increase the tail distribution~\cite{perez2017effectiveness}. Bi-lateral Branch Networks (BBNs)~\cite{zhou2020bbn} use two network branches, a conventional learning branch and a re-balancing branch, to address the long-tailed recognition problem~\cite{zhou2020bbn}. Learning from Multiple Experts (LFME)~\cite{xiang2020learning} and Routing Diverse Experts (RIDE)~\cite{wang2020long} adopt the same idea of ensemble learning to aggregate knowledge from multiple experts. 

Recently, a two-step training method decoupled the representation learning and classifier learning, achieving superior performance compared to previous methods~\cite{kang2019decoupling}. Mixup Shifted Label-Aware Smoothing (MiSLAS)~\cite{zhong2021improving} uses a regularization technique mixup~\cite{zhang2017mixup} to further improve in a two-stage framework. Different from these works addressing the long-tailed learning problem in a static world, we propose class incremental learning scenarios with long-tailed distributions. This requires continually learning different long-tailed classes in a dynamic world without catastrophic forgetting.

%% file: method.tex
In this section we first formulate two long-tailed CIL scenarios and then we adapt several existing state-of-the-art methods for conventional CIL to long-tailed CIL scenarios. Finally, we propose a two-stage training method with a learnable weight scaling layer for long-tailed CIL.

\subsection{Long-tailed CIL}
In conventional CIL the model must sequentially learn different tasks where each new task consists of a set of new classes. Formally, for each training task $t$, the data is denoted as $\mathcal{D}_t$, where  $\mathcal{D}_t = {(\mathbf{x}_t^{(i)}, y_t^{(i)})}_{i=1}^{n_t}$, and $\mathbf{x}_t^{(i)}$ is an input image, $y_t^{(i)}$ is the corresponding label and there are $n_t$ samples in total at task $t$. Normally the number of samples per class is equally distributed and it can be calculated as $\frac{n_t}{C_t}$, in which $C_t$ is the number of classes at task $t$. Therefore, the total number of classes learned up to current tasks is $C_{1:t}$. For replay-based methods, at each training task a memory of $\mathcal{M}$ (known as the memory of exemplars)is stored for previous classes up to task $t-1$, normally $\left \lfloor \frac{ |\mathcal{M}| }{C_{1:t-1}} \right \rfloor$ samples stored for each class.

\begin{figure}[t]
\begin{center}
\includegraphics[width=0.5\linewidth]{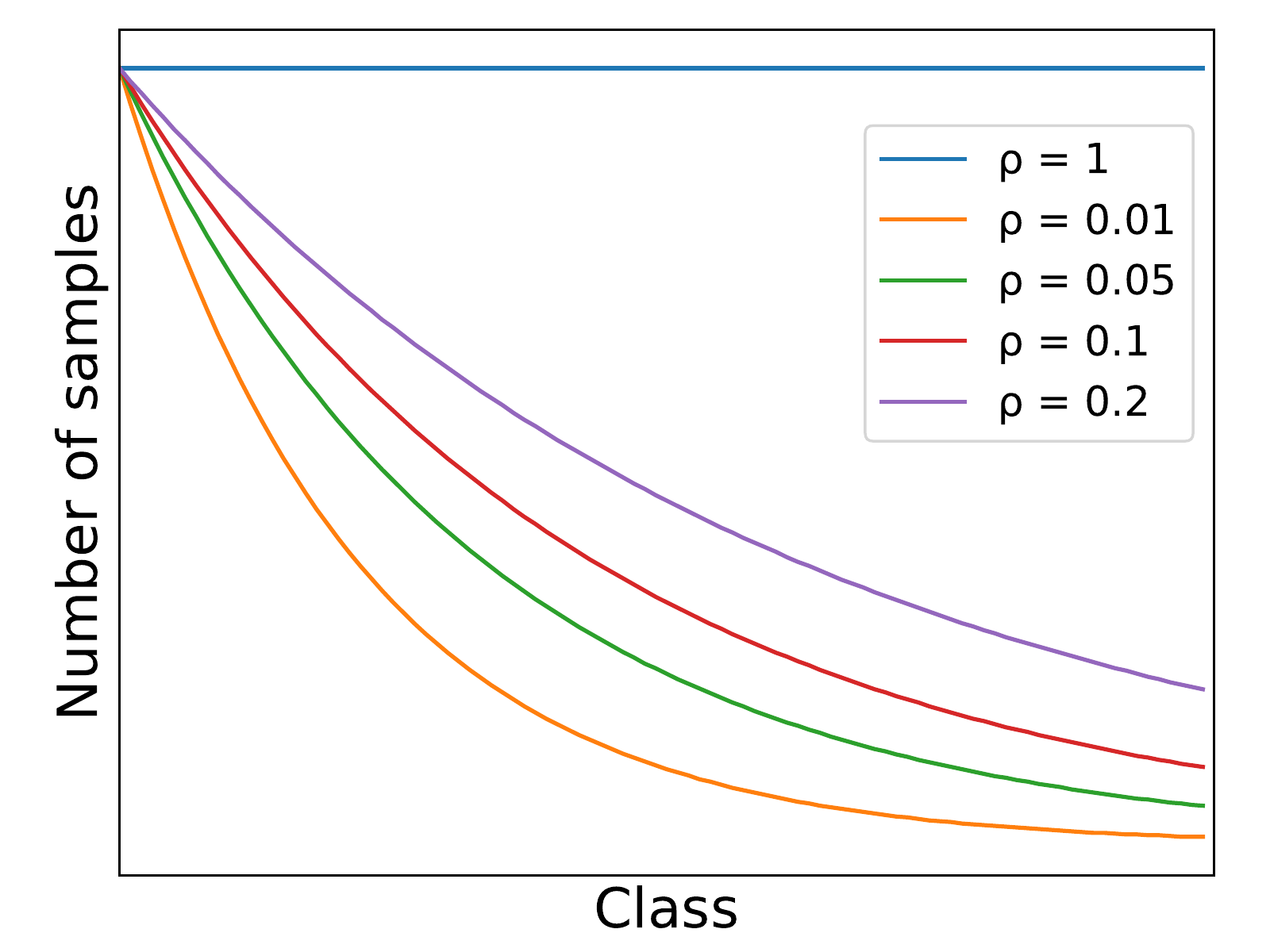}
\end{center}
\caption{ An illustration imbalance ratios $\rho$ for long-tailed distribution generation. $\rho=1$ is the balanced distribution and corresponds to conventional CIL.
}
\label{fig:ratio}
\end{figure}

While class-incremental learning has many practical applications, the assumption of equally distributed samples is not always realistic. Most real-world class distributions are in face \emph{long-tailed}. The long-tailed distribution follows an exponential decay in sample sizes across classes as described in~\cite{cao2019learning}. This decay is parameterized by $\rho$ which is the ratio between the most and least frequent classes. An example of different ratios can be found in Fig.~\ref{fig:ratio}, the larger the ratio, the more balanced the distribution. $\rho=1$ is the conventional CIL case and $\rho$ in (0,1) indicates different degrees of long-tailed distribution. 

Given a sampled long-tailed class distribution, we propose two long-tailed CIL scenarios constructed from it:
\begin{itemize}
\item \textbf{Ordered LT-CIL} which starts learning from the most frequent classes as first task and ends with the last task containing the least frequent classes; and
\item \textbf{Shuffled LT-CIL} which first shuffles the long-tailed distribution randomly, and then constructs different tasks based on the Shuffled class order. It thus has varying degrees of imbalance in each task.
\end{itemize}
In both conventional and long-tailed CIL the test set contains uniformly distributed samples for each class. Ordered LT-CIL is representative of real world applications where we are able to learn from easy-to-sample classes first, and then gradually increase the difficulty of learning with less frequent samples. Shuffled LT-CIL, on the other hand, considers a more general scenario without any assumptions of the arriving data distribution.   

\begin{figure*}[t]
\includegraphics[width=0.95\textwidth]{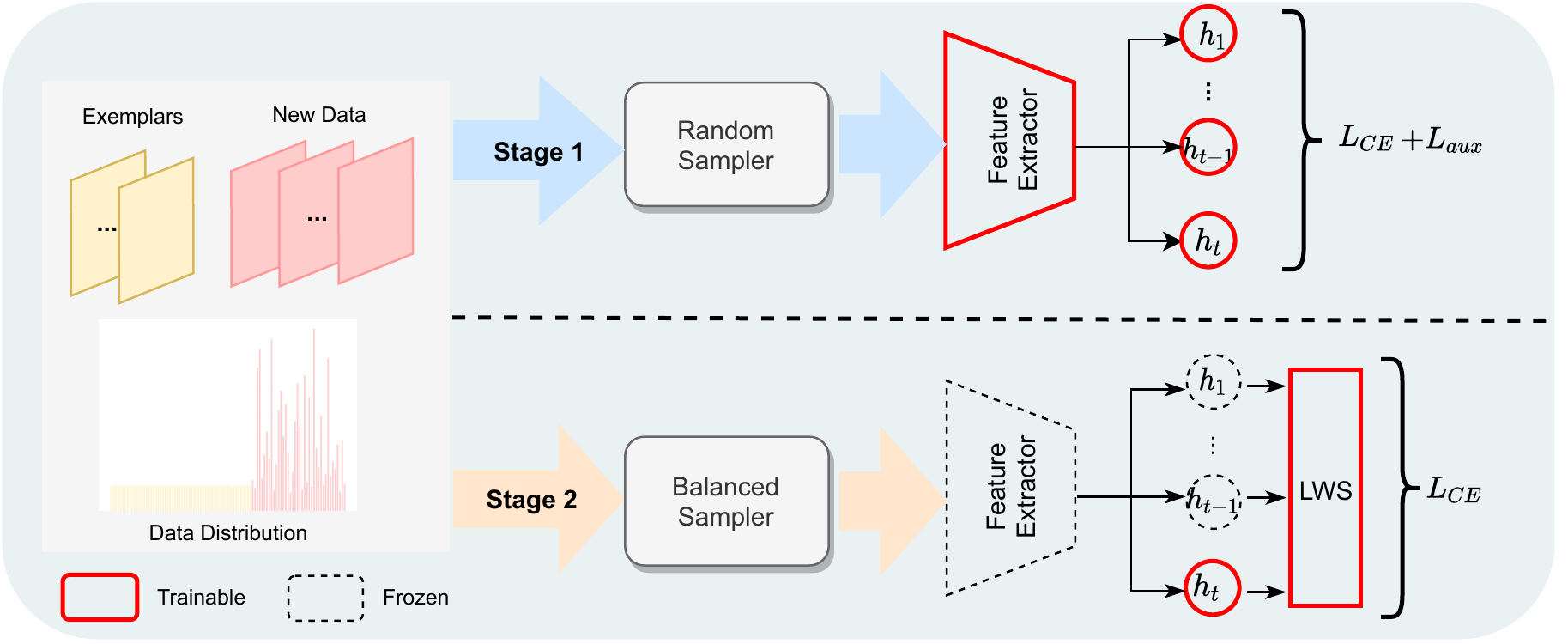}
\caption{An overview of our two-stage method with a learnable weight scaling layer for LT-CIL. Note that new data for the current task follows a long-tailed distribution and the memory contains a few samples from previous tasks. In the first stage, random sampling is used to learn a better feature extractor together with $L_{\text{CE}}$ and method-specific loss $L_{\text{aux}}$ to reduce forgetting.In the second stage, a balanced sampler is used to learn a balanced classifier together with a layer of learnable scaling weights (LWS). To reduce representation drift for future tasks, we fix the previous classifiers $h_{1:t-1}$ and only train the current head and LWS with cross entropy loss $L_{CE}$. 
}
\label{fig:overview}
\end{figure*}

\subsection{Conventional CIL methods applied to LT-CIL}
A classification model can be roughly divided into two parts: a feature extractor $f_\theta$, usually a convolutional neural network with parameters $\theta$, and a classification head $h_\phi$ with parameters $\phi$. The cross entropy loss $\mathcal{L}_{CE}$ at task $t$ is defined as:
\begin{equation}
\label{eq:crossentropy}
    \mathcal{L}_{\text{CE},t} (\mathbf{x}, y; \theta_{t}, \phi_{t}) = - \frac{1}{|\mathcal{D}_t|+|\mathcal{M}|} \sum_{(\mathbf{x}, y) \in \mathcal{D}_{t}\cup \mathcal{M} }\mathbf{y}\cdot \text{log}( \mathbf{\mathbf{p}}_{1:t}(\mathbf{x})),
\end{equation}
where $\mathbf{y}$ is a one-hot vector with 1 at the position of the correct class, and $\mathbf{p}_{1:t}(\mathbf{x})$ is a vector containing the probability predictions for image $\mathbf{x}$ over all classes up to task $t$.

Directly minimizing Eq.~\ref{eq:crossentropy}, even if replaying past-task exemplars sampled from $\mathcal{M}$, can result in catastrophic forgetting. Normally, a method-specific auxiliary loss $L_{\text{aux}}$ is added to mitigate forgetting using regularization or by replaying past-task exemplars from $\mathcal{M}$. Examples of such an auxiliary loss include the many techniques based on knowledge distillation~\cite{douillard2020podnet,hou2019learning,li2018learning}.The total loss $L_{t}$ for task $t$ is thus:
\begin{equation}
    L_{t} = L_{\text{CE},t} + L_{\text{aux},t}
    \label{eq:stage1}
\end{equation}
where $L_{\text{aux}}$ is the method-specific loss.
In Section~\ref{sec:exp} we evaluate the LwF~\cite{li2018learning}, EEIL~\cite{castro2018end}, LUCIR~\cite{hou2019learning} and PODNET~\cite{douillard2020podnet} methods on our Ordered and Shuffled LT-CIL scenarios. These methods can be directly applied to both Ordered LT-CIL and Shuffled LT-CIL scenarios, but since they are not specifically designed for LT-CIL we are interested in how they perform in these more challenging scenarios.

\subsection{A two-stage method with a learnable weight scaling layer}
Two-stage methods have shown state-of-the-art performance in long-tailed recognition~\cite{chu2020feature,kang2019decoupling,zhang2021distribution,zhong2021improving}. In general, two-stage learning decouples representation learning from classifier learning:
\begin{itemize}
    \item In the \textbf{first stage}, it aims to learn a better feature extractor $f_\theta$ using an instance-balanced sampler (also known as random sampler where each data point has the \textit{same probability} of being sampled) that generalizes well; 
    \item In the \textbf{second stage}, a class-balanced sampler in which \emph{each class} is sampled uniformly first and then each instance is sampled uniformly within it (also known as balanced sampler) is used to retrain the classifier $h_\phi$ to obtain better classification accuracy.
\end{itemize}

In LT-CIL, for task $t$ we first learn a model from the first stage (seen in the top part of Fig.~\ref{fig:overview}) using Eq.~\ref{eq:stage1} with both cross entropy loss $L_{CE,t}$ and method-specific loss $L_{aux,t}$.
In the second stage, we attach a single trainable layer, which we call a Learnable Weight Scaling (LWS) layer, $\mathbf{W}\in\mathrm{R}^{C_{1:t}\times 1}$ with dimension equal to the number of classes $C_{1:t}$ in the output of classifier $h_{1:t}$ at task $t$ (as shown in the bottom part of Fig.~\ref{fig:overview}).

The LWS is used to balance between classes with different numbers of samples in the long-tailed distribution. The final output of the model $\hat{\mathbf{z}}$ is calculated using an element-wise product of the classifier output with the LWS:
\begin{equation}
    \hat{\mathbf{z}} = \mathbf{W} \odot h_{\phi_{1:t}}(f_{\theta_{t}}(\mathbf{x}))
\end{equation}

We found it to be essential to fix previous head $h_{\phi_{1:t-1}}$ at the second stage when learning together with LWS layer, otherwise the modified $h_{\phi_{1:t-1}}$ can back-propagate in the future tasks and damage representation learning in the first stage. Note that only $L_{\text{CE}}$ loss is used in the second stage  no matter which loss is chosen for $L_{\text{aux}}$, and that the LWS layer $\mathbf{W}$ is applied only in the second stage for training and evaluation. The scaling layer for task $t$ is discarded in the first stage training for the next task $t+1$.

\paragraph{Discussion.} Conventional long-tailed learning considers a fixed set of classes,
therefore it is challenging to apply two-stage methods directly to incremental learning for LT-CIL. In the first stage, a good representation must be learned and catastrophic forgetting avoided. Whereas in the second stage, the classifier learned from balanced sampling will be the initialization for the future tasks. Thus, the modified classifier can lead to representation drift back-propagated to future tasks, which harms the generalization of feature representation in the dynamic and incremental process of continual learning.

%% file: experiments.tex
\newcommand\mainaccuracy[1]{{{}}{#1}}
\newcommand\plusaccuracy[1]{{{}}{\footnotesize{\color{red}#1}}}

In this section we first introduce the experimental setup and then compare different existing CIL methods applied to LT-CIL benchmarks. Finally we evaluate our two-stage method and conduct ablation study of key elements. 

\subsection{Experimental setup}
\paragraph{Implementation details.} We experiment on two datasets: CIFAR-100 and ImageNet-Subset with 100 classes.
We use the publicly available implementations of existing CIL methods in the  framework FACIL~\cite{masana2020class} and implement our two-stage algorithm with long-tailed data loader in the same framework for fair comparison. We follow LUCIR and PODNET by starting with a large first task with half of the classes in each dataset and equally dividing the remaining classes in subsequent tasks.

We use ResNet-32 for CIFAR-100 ResNet-18 for ImageNet-Subset. We use an initial learning rate of 0.1, and divide it by 10 after 80 and 120 epochs (160 epochs in total) for CIFAR-100. ImageNet-Subset, the learning rate starts from 0.1 and is divided by 10 after 30 and 60 epochs (90 epochs in total). The batch size is 128 for all experiments. For stage two training,  the learning rate is set to 0.1 and we train it for 30 epochs.

\paragraph{Evaluation protocols.} We use average accuracy over all classes and average incremental accuracy over all tasks as evaluation metrics. We first evaluate different methods on LT-CIL scenarios with an imbalance ratio $\rho=0.01$ and 20 exemplars per class, and report varying imbalance ratios and number of exemplars in Section~\ref{sec:ablation}.

\subsection{Conventional methods on LT-CIL scenarios}
\begin{figure*}[t]
\begin{minipage}[b]{0.32\linewidth}
\centering
\includegraphics[width=\textwidth]{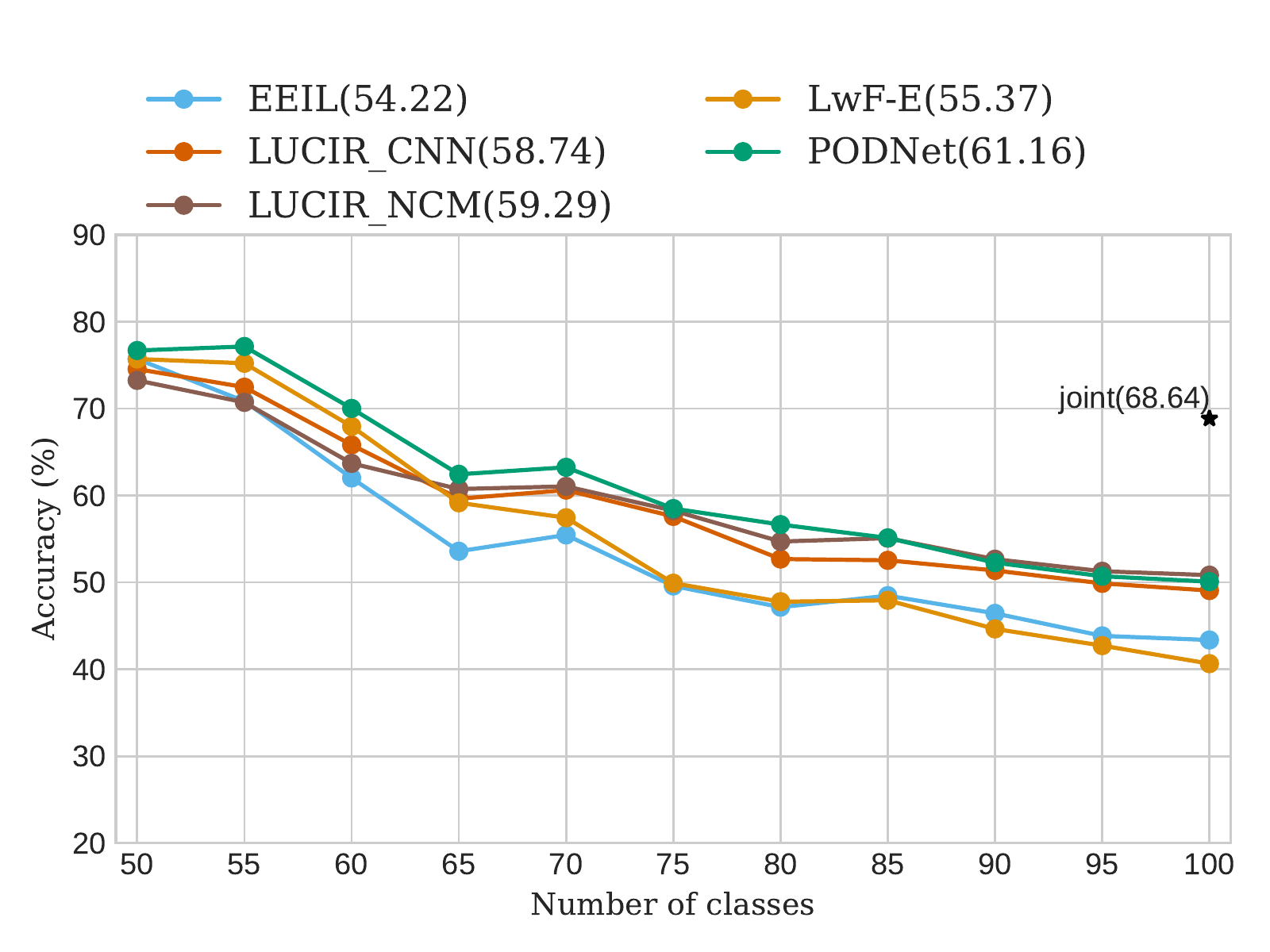}
\subcaption{Conventional CIL}
\end{minipage}
\begin{minipage}[b]{0.32\linewidth}
\centering
\includegraphics[width=\textwidth]{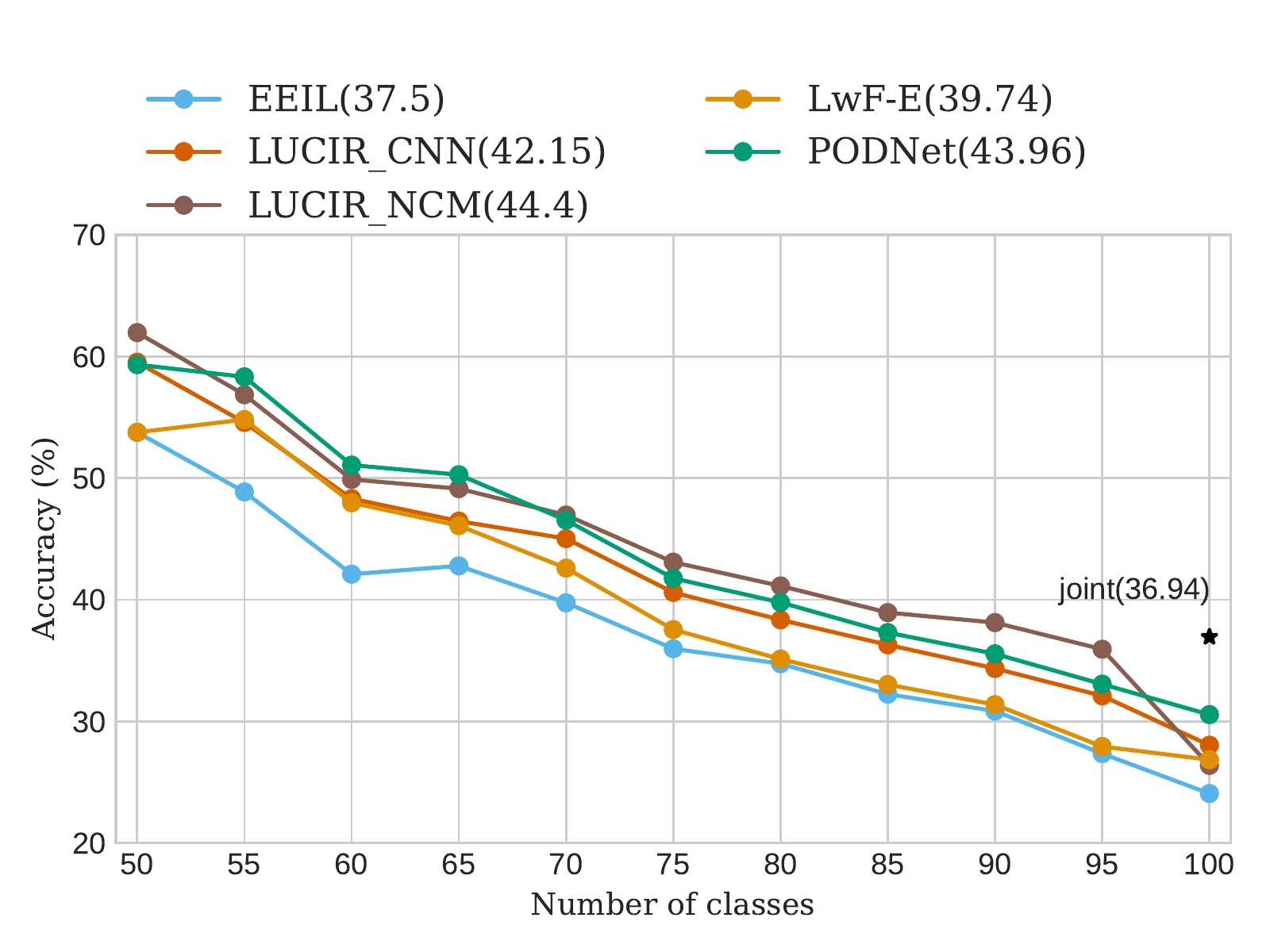}
\subcaption{Ordered LT-CIL}
\end{minipage}
\begin{minipage}[b]{0.32\linewidth}
\centering
\includegraphics[width=\textwidth]{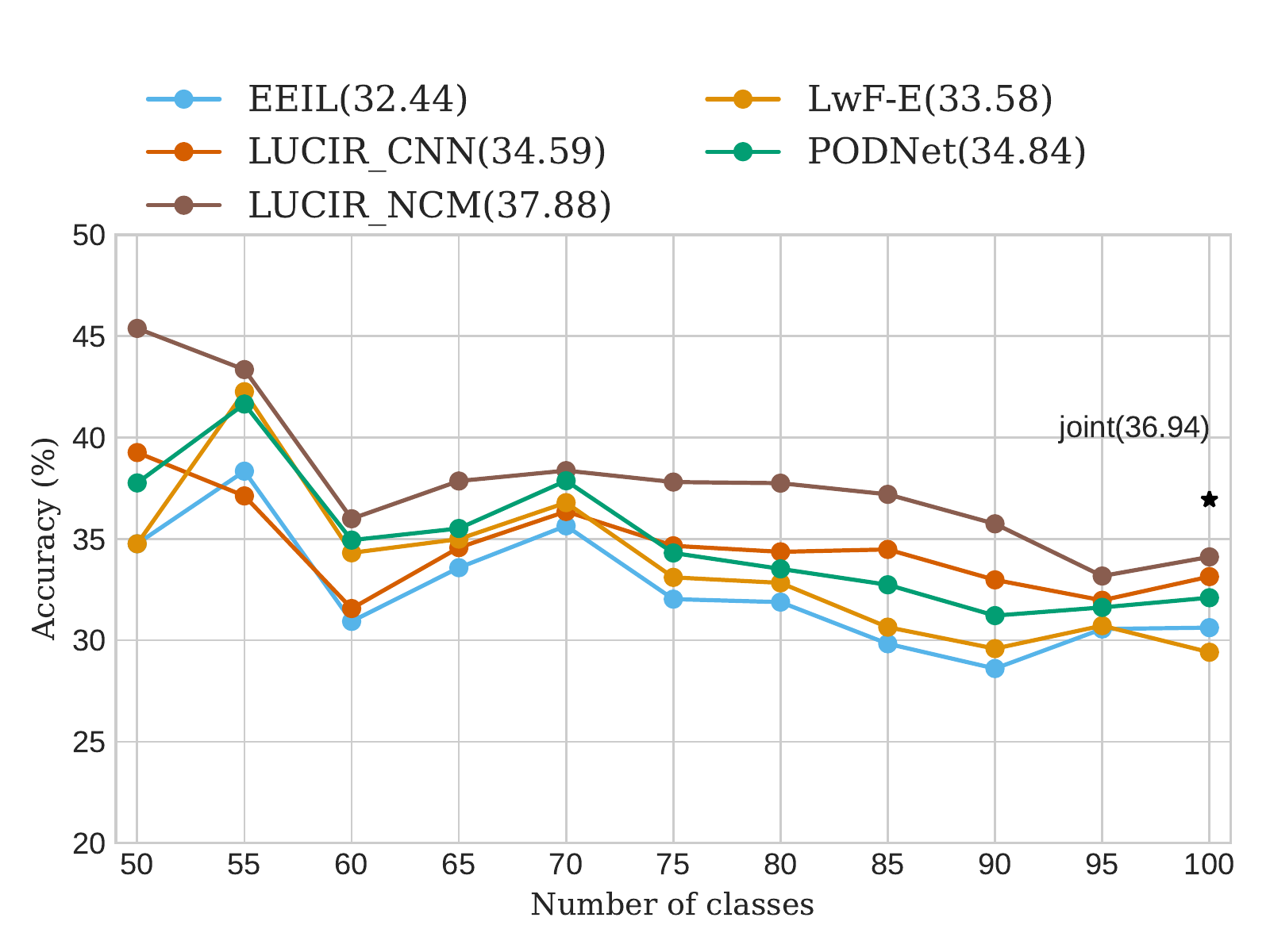}
\subcaption{Shuffled LT-CIL}
\end{minipage}
\caption{Average accuracy for different scenarios on CIFAR-100. Average incremental accuracy is in the parentheses.
}
\label{fig:cifar}
\end{figure*}

In this section we analyze the performance of four popular CIL methods: LwF~\cite{li2018learning} with exempalrs, EEIL~\cite{castro2018end}, LUCIR~\cite{hou2019learning}, and PODNET~\cite{douillard2020podnet}). 
In Fig.~\ref{fig:cifar}(a) we first evaluate on conventional CIL as a reference. We then evaluate on the Ordered LT-CIL setting in Fig.~\ref{fig:cifar}(b). It is clear that LT-CIL is a more challenging scenario given that joint training drops from 68.64 to 36.94. Interestingly, LUCIR with nearest class mean (NCM) classifier obtains the best performance in average incremental accuracy. LUCIR is much better than PODNET for the tail classes with few samples in the end of training, except for the last task.

Similarly, as seen in Fig.~\ref{fig:cifar}(c), the overall performance on Shuffled LT-CIL scenario for all methods is much worse than in the conventional CIL scenario. LUCIR with NCM classifier again achieves the best performance.

\begin{figure*}[t]
\begin{minipage}[b]{0.32\linewidth}
\centering
\includegraphics[width=\textwidth]{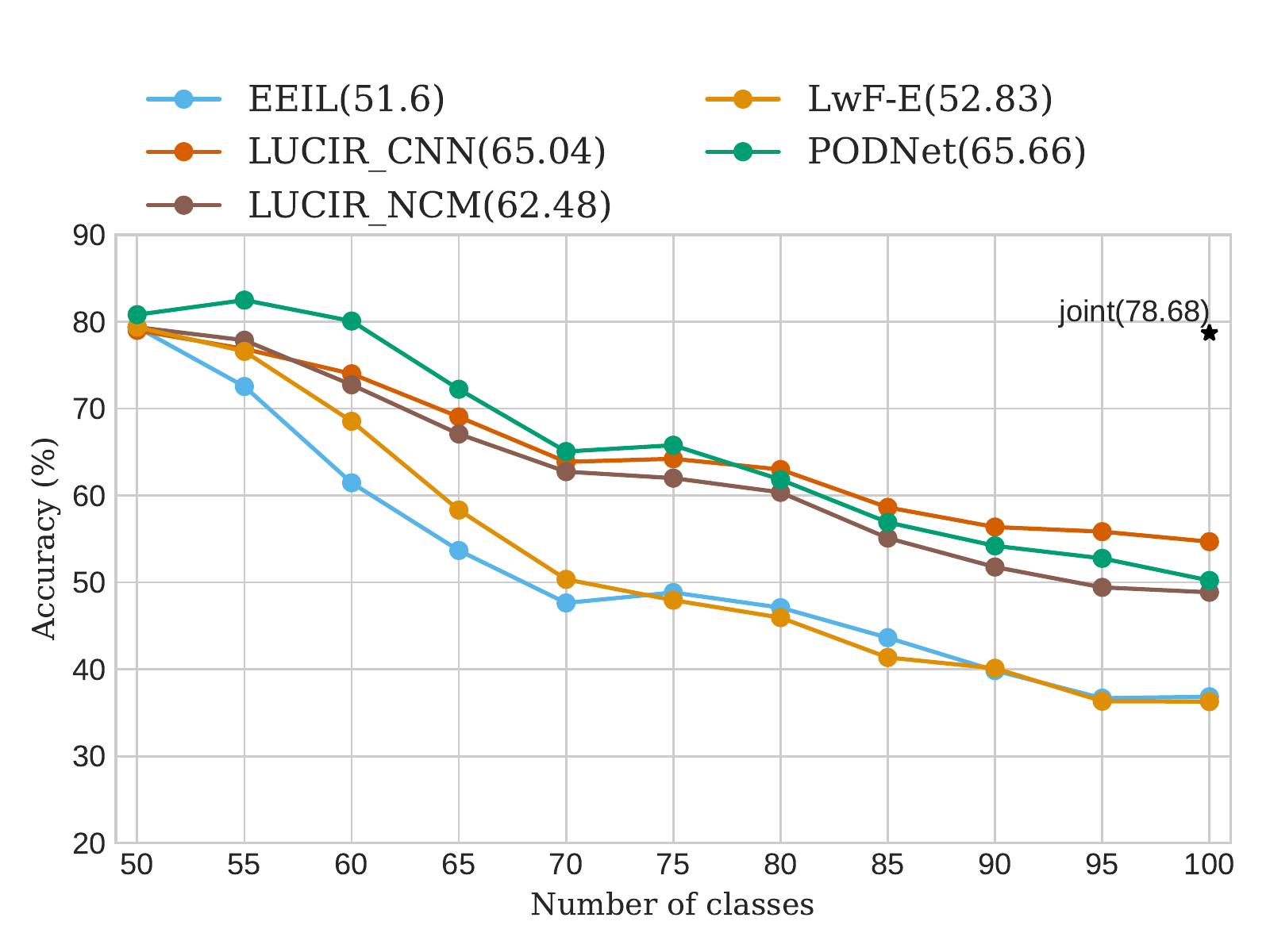}
\subcaption{Conventional CIL}
\end{minipage}
\begin{minipage}[b]{0.32\linewidth}
\centering
\includegraphics[width=\textwidth]{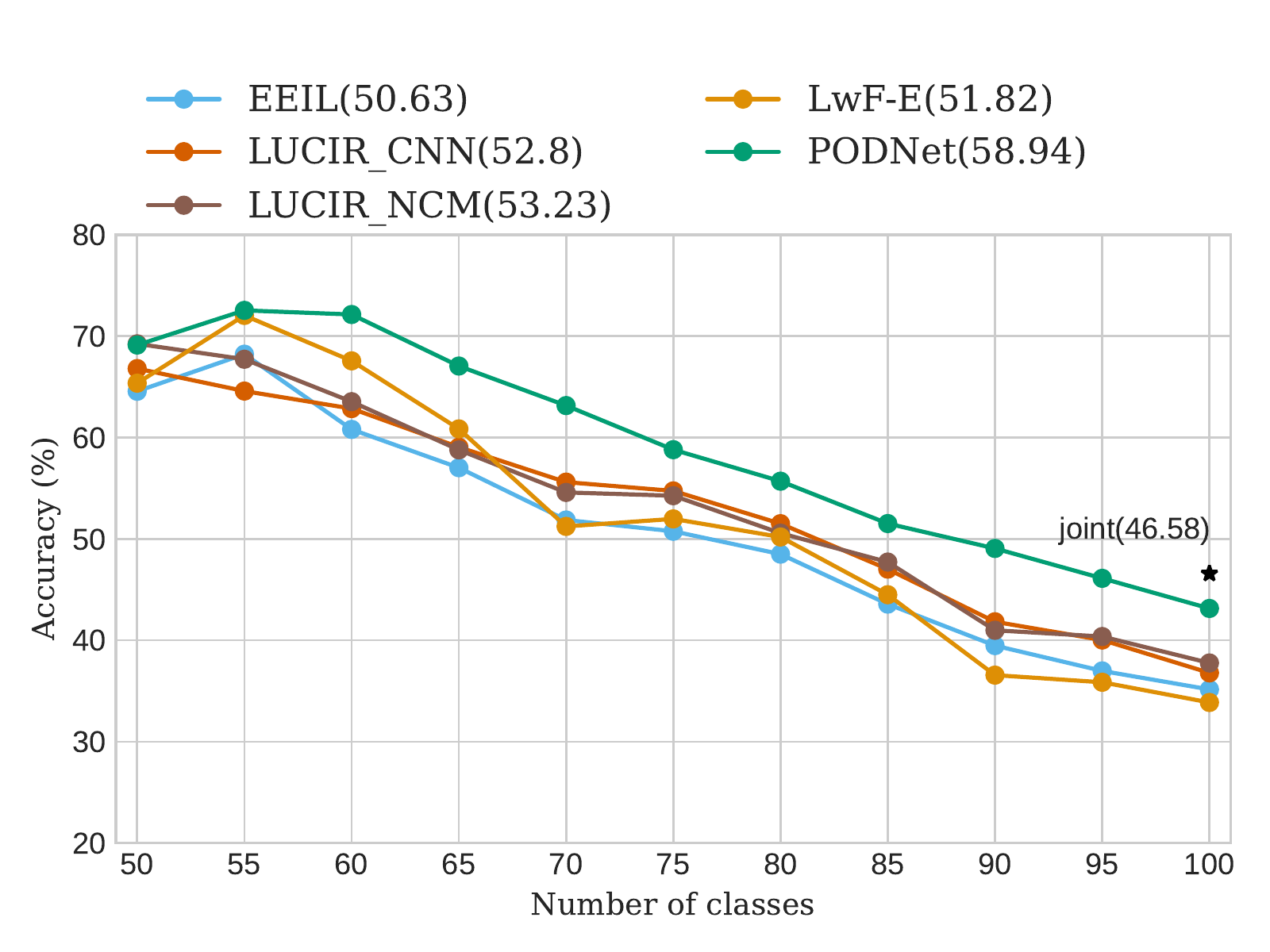}
\subcaption{Ordered LT-CIL}
\end{minipage}
\begin{minipage}[b]{0.32\linewidth}
\centering
\includegraphics[width=\textwidth]{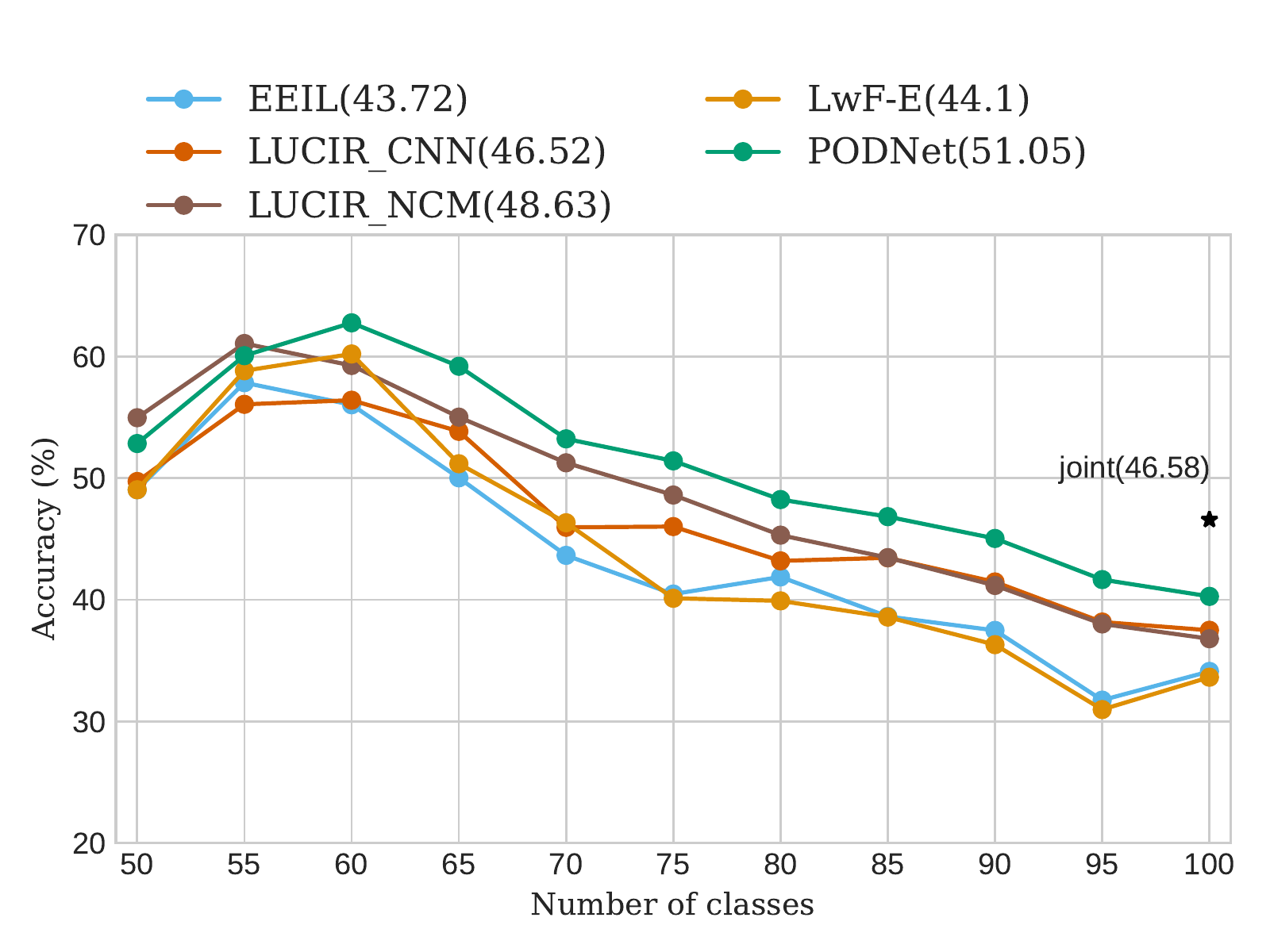}
\subcaption{Shuffled LT-CIL}
\end{minipage}
\caption{Average accuracy for different scenarios on ImageNet-Subset. Average incremental accuracy is in parentheses.
}
\label{fig:imagenet}
\end{figure*}

In Fig.~\ref{fig:imagenet} we report on the same experiment performed on ImageNet-Subset. When we apply these methods to LT scenarios, PODNET achieves significantly better accuracy compared to other methods with average incremental  accuracy of 58.94 and 51.05 for Ordered LT-CIL and Shuffled LT-CIL, respectively. For conventional CIL, methods often has similar rankings in terms of performance for different datasets as shown in Fig.~\ref{fig:cifar} (a) and ~\ref{fig:imagenet} (a). While for LT-CIL, we can see that LUCIR outperforms PODNET on CIFAR-100 but achieves worse results than PODNET on ImageNet-Subset. It suggests that these methods are not as robust as in the conventional CIL scenario.

\subsection{Results for our two-stage method}

\begin{table*}[t]
\setlength{\tabcolsep}{6mm}{
  \centering
  \resizebox{0.99\columnwidth}{!}{%
  \begin{tabular}{ll*{2}{r@{}l}c*{2}{r@{}l}c*{2}{r@{}l}}
  \toprule
  \multicolumn{2}{c}{} & \multicolumn{4}{c}{\textbf{CIFAR-100}} & \textbf{} & \multicolumn{4}{c}{\textbf{ImageNet-Subset}} \\ \cline{3-6} \cline{8-11} \cline{13-16} 
  \multicolumn{2}{l}{\multirow{-2}{*}{\quad \quad \textbf{Methods}}} & \multicolumn{2}{l}{\textit{5 tasks}} & \multicolumn{2}{l}{\textit{10 tasks}} & \textit{} & \multicolumn{2}{l}{\textit{5 tasks}} & \multicolumn{2}{l}{\textit{10 tasks}} & \textit{} \\ \hline
  \multirow{6}{*}{\rotatebox{90}{\textit{Ordered LT-CIL}}}
    & EEIL   & 38.46 &  & 37.50 &  &  & 50.68 &  & 50.63 & &  \\
  & + (Ours) & \mainaccuracy{38.97} & \plusaccuracy{+0.51}&\mainaccuracy{37.58} & \plusaccuracy{+0.08} &\mainaccuracy{} &\mainaccuracy{51.36} & \plusaccuracy{+0.68} &\mainaccuracy{50.74} & \plusaccuracy{+0.11} \\ \cline{2-16} 
  & LUCIR   & 42.69 &  & 42.15 &  &  & 52.91 &  & 52.80 & &  \\
  & + (Ours) & \textbf{\mainaccuracy{45.88}} & \plusaccuracy{+3.19}&\textbf{\mainaccuracy{45.73}} & \plusaccuracy{+3.58} &\mainaccuracy{} &\mainaccuracy{54.22} & \plusaccuracy{+1.31} &\mainaccuracy{55.41} & \plusaccuracy{+2.61} \\ \cline{2-16} 
& PODNET   & 44.07 &  & 43.96 &  &  & 58.78 &  & 58.94 & &  \\
  & + (Ours) & \mainaccuracy{44.38} & \plusaccuracy{+0.31}&\mainaccuracy{44.35} & \plusaccuracy{+0.39} &\mainaccuracy{} &\textbf{\mainaccuracy{58.82}} & \plusaccuracy{+0.04} &\textbf{\mainaccuracy{59.09}} & \plusaccuracy{+0.15} \\ \cline{2-16} 
  \hline
  \multirow{6}{*}{\rotatebox{90}{\textit{Shuffled LT-CIL}}}
  & EEIL   & 31.91 &  & 32.44 &  &  & 42.87 &  & 43.72 & &  \\
  & + (Ours) & \mainaccuracy{34.19} & \plusaccuracy{+2.28}&\mainaccuracy{33.70} & \plusaccuracy{+1.26} &\mainaccuracy{} &\mainaccuracy{49.31} & \plusaccuracy{+6.44} &\mainaccuracy{48.26} & \plusaccuracy{+4.54} \\ \cline{2-16} 
  & LUCIR   & 35.09 &  & 34.59 &  &  & 45.80 &  & 46.52 & &  \\
  & + (Ours) & \textbf{\mainaccuracy{39.40}} & \plusaccuracy{+4.31}&\textbf{\mainaccuracy{39.00}} & \plusaccuracy{+4.41} &\mainaccuracy{} &\textbf{\mainaccuracy{52.08}} & \plusaccuracy{+6.28} &\mainaccuracy{51.91} & \plusaccuracy{+5.39} \\ \cline{2-16} 
  & PODNET   & 34.64 &  & 34.84 &  &  & 49.69 &  & 51.05 & &  \\
  & + (Ours) & \mainaccuracy{36.37} & \plusaccuracy{+1.73}&\mainaccuracy{37.03} & \plusaccuracy{+2.19} &\mainaccuracy{} &\mainaccuracy{51.55} & \plusaccuracy{+1.86} &\textbf{\mainaccuracy{52.60}} & \plusaccuracy{+1.55} \\ \cline{2-16} 
  \hline\hline
  \multirow{6}{*}{\rotatebox{90}{\textit{Conventional CIL}}}
   & EEIL   & 57.41 &  & 54.22 &  &  & 53.84 &  & 47.30 & &  \\
  & + (Ours) & \mainaccuracy{59.10} & \plusaccuracy{+1.69}&\mainaccuracy{56.91} & \plusaccuracy{+2.69} &\mainaccuracy{} &\mainaccuracy{57.45} & \plusaccuracy{+3.61} &\mainaccuracy{53.40} & \plusaccuracy{+6.10} \\ \cline{2-16} 
  & LUCIR   & 61.15 &  & 58.74 &  &  & 67.21 &  & 65.04 & &  \\
  & + (Ours) & \mainaccuracy{63.48} & \plusaccuracy{+2.33}&\mainaccuracy{60.57} & \plusaccuracy{+1.83} &\mainaccuracy{} &\mainaccuracy{68.82} & \plusaccuracy{+1.61} &\mainaccuracy{67.44} & \plusaccuracy{+2.40} \\ \cline{2-16} 
  & PODNET   & 63.15 &  & 61.16 &  &  & 70.13 &  & 65.66 & &  \\
  & + (Ours) & \textbf{\mainaccuracy{64.58}} & \plusaccuracy{+1.43}&\textbf{\mainaccuracy{62.63}} & \plusaccuracy{+1.47} &\mainaccuracy{} &\textbf{\mainaccuracy{71.08}} & \plusaccuracy{+0.95} &\textbf{\mainaccuracy{68.47}} & \plusaccuracy{+2.81} \\  

\addlinespace
  \bottomrule
  \end{tabular}}
  \caption{Comparison of average incremental accuracy on CIFAR-100 and ImageNet-Subset in the LT-CIL and conventional CIL scenarios.}
  \label{tab:sota}
  }
\end{table*}

\paragraph{Our method for LT-CIL.} In Table~\ref{tab:sota} we integrate our proposed two-stage strategy into three existing methods: EEIL, LUCIR (with CNN classifier), and PODNET. In general, the two-stage strategy helps on all three methods in both 5- and 10-task settings. The improvement is especially noticeable in the Shuffled LT-CIL scenario. Specifically, for EEIL, our method only improves by a small margin on Ordered LT-CIL scenario but boosts significantly on Shuffled LT-CIL. It outperforms EEIL by 2.28 and 1.26 on CIFAR-100 when $T=5$ and $T=10$, respectively for Shuffled LT-CIL. The improvement is even larger on ImageNet-Subset with 6.44 and 4.54 improvement in absolute accuracy. For LUCIR, we see a consistent boost by adding our method, improving from 1.31 to 6.28 for CIFAR-100 and ImageNet-Subset, respectively. PODNET is the best baseline in most scenarios where we observe a smaller gain with our proposed method compared to LUCIR. Overall, PODNET and LUCIR with our method can achieve very competitive results, which improves the consistency for both Ordered LT-CIL and Shuffled LT-CIL.

\paragraph{Our method for conventional CIL.}
Surprisingly, as seen in Table~\ref{tab:sota}, when we combine ours with existing methods the performance is improved not only in LT-CIL scenarios but also for conventional CIL. We believe this is due to the imbalance caused by limited memory for storing exemplars from previous tasks.

\paragraph{Results on real-world long-tailed dataset} We experiment with 100 classes chosen from the iNaturalist dataset. We randomly chose 100 classes from the pantae super category and tested LUCIR and LUCIR$+$ with the data seperated into 5 tasks with a base task of 50 classes. Results show that LUCIR can achieve an accuracy of 32.34$\%$, and LUCIR+ with two stage training about 1.46$\%$ higher. iNaturalist is a real-world dataset with long-tailed distribution,
and thus the value of $\rho$ is undefined. We estimate it to be
about 0.01. It shows how our method perform in real-world
dataset under long-tailed distribution.

\paragraph{Further results on AANets~\cite{liu2021adaptive} and DER~\cite{yan2021dynamically}} We report results for
AANets (based on LUCIR) on Shuffle LT-CIL scenario (CIFAR-100 with 10-task setting). AANets outperforms LUCIR by a large margin achieving 38.53 in average incremental accuracy, and adding our method still improves over it by
about 1\%. We found that DER does not
work well on long-tail scenarios ( with only 29.54 in average accuracy), but our method
improves it by about 4\%.

\begin{figure*}
\begin{minipage}[b]{0.32\linewidth}
\centering
\includegraphics[width=\textwidth]{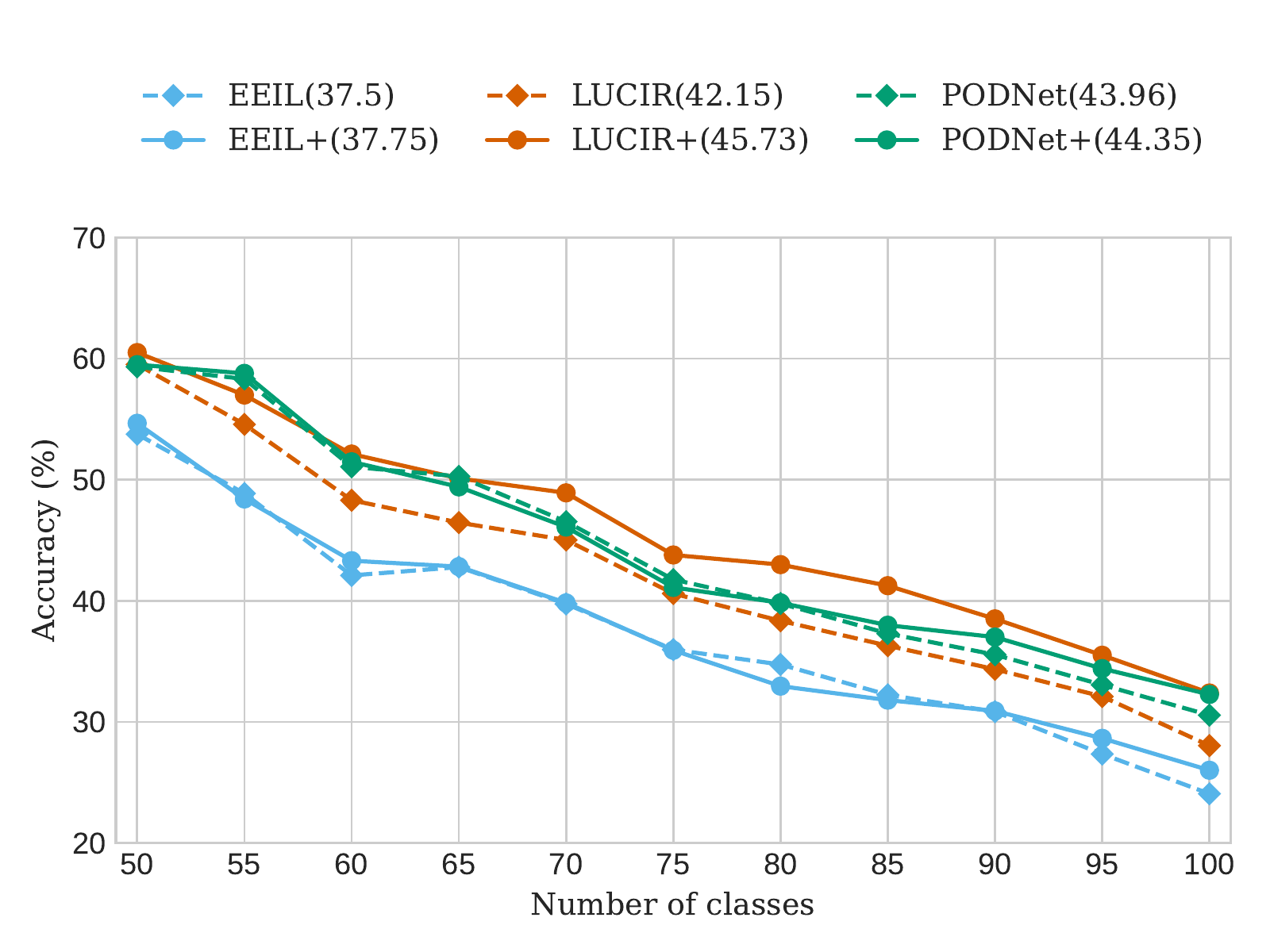}
\subcaption{$\rho$ = 0.01 Ordered LT}
\end{minipage}
\begin{minipage}[b]{0.32\linewidth}
\centering
\includegraphics[width=\textwidth]{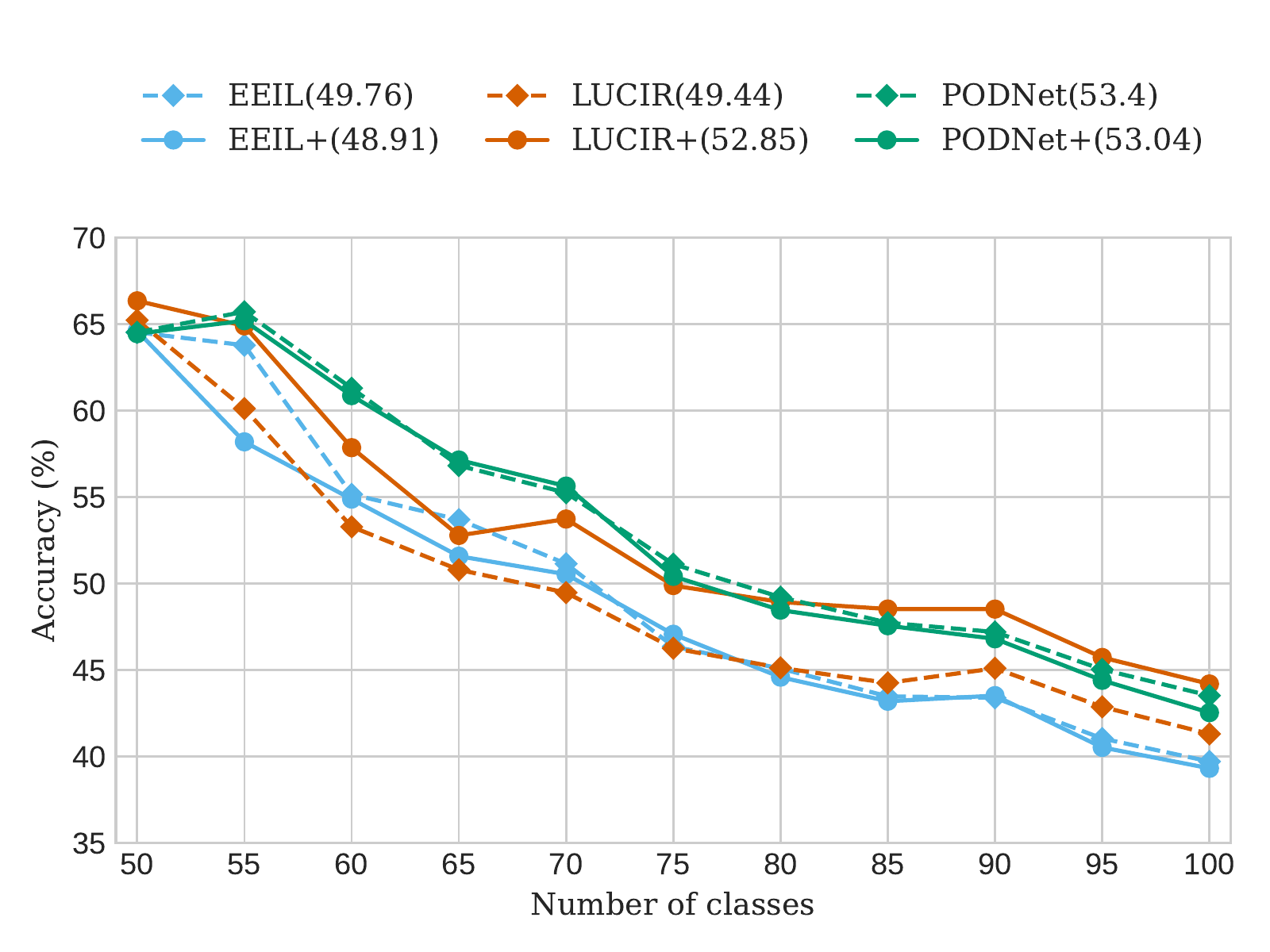}
\subcaption{$\rho$ = 0.05 Ordered LT}
\end{minipage}
\begin{minipage}[b]{0.32\linewidth}
\centering
\includegraphics[width=\textwidth]{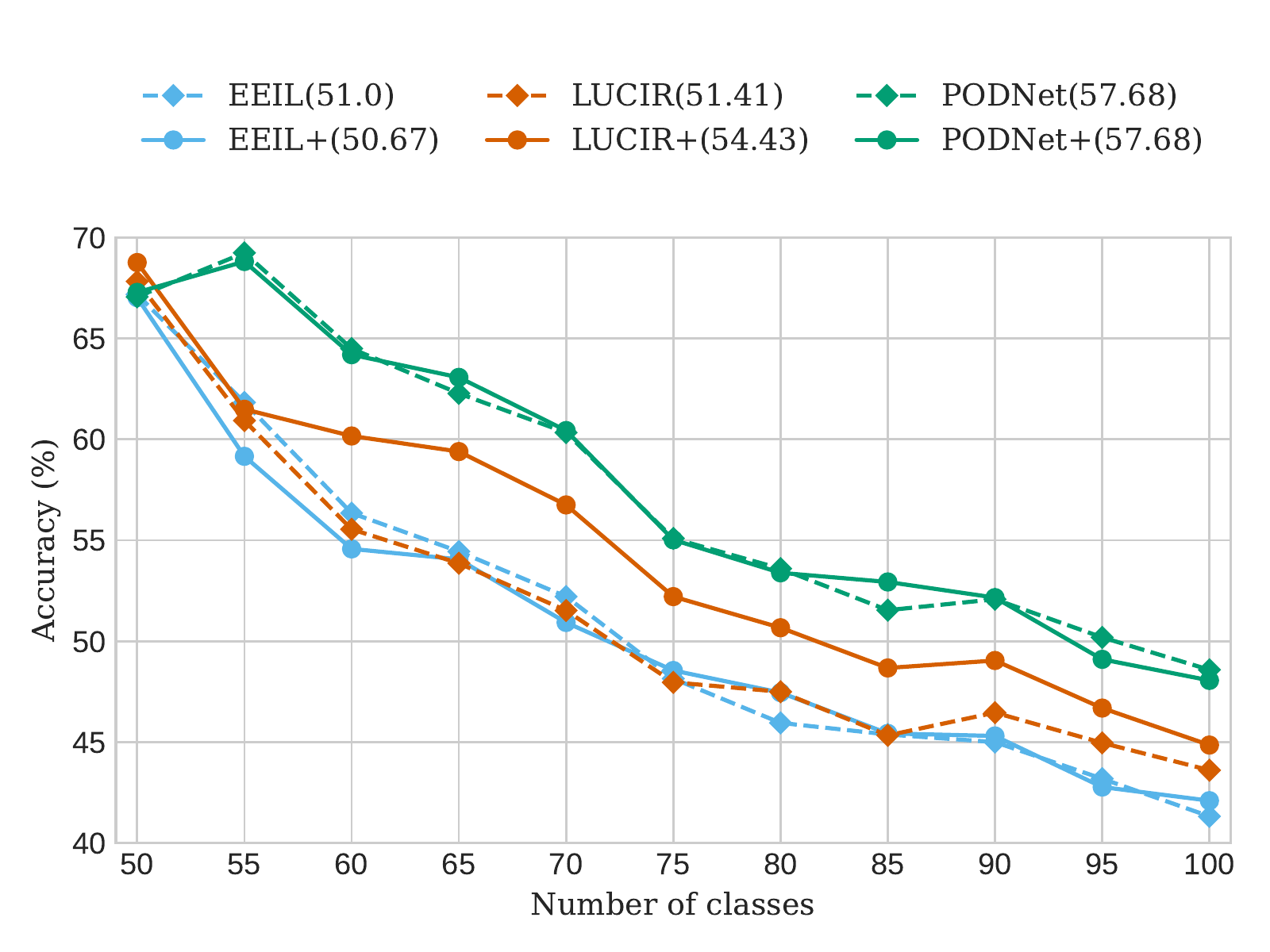}
\subcaption{$\rho$ = 0.1 Ordered LT}
\end{minipage}
\begin{minipage}[b]{0.32\linewidth}
\centering
\includegraphics[width=\textwidth]{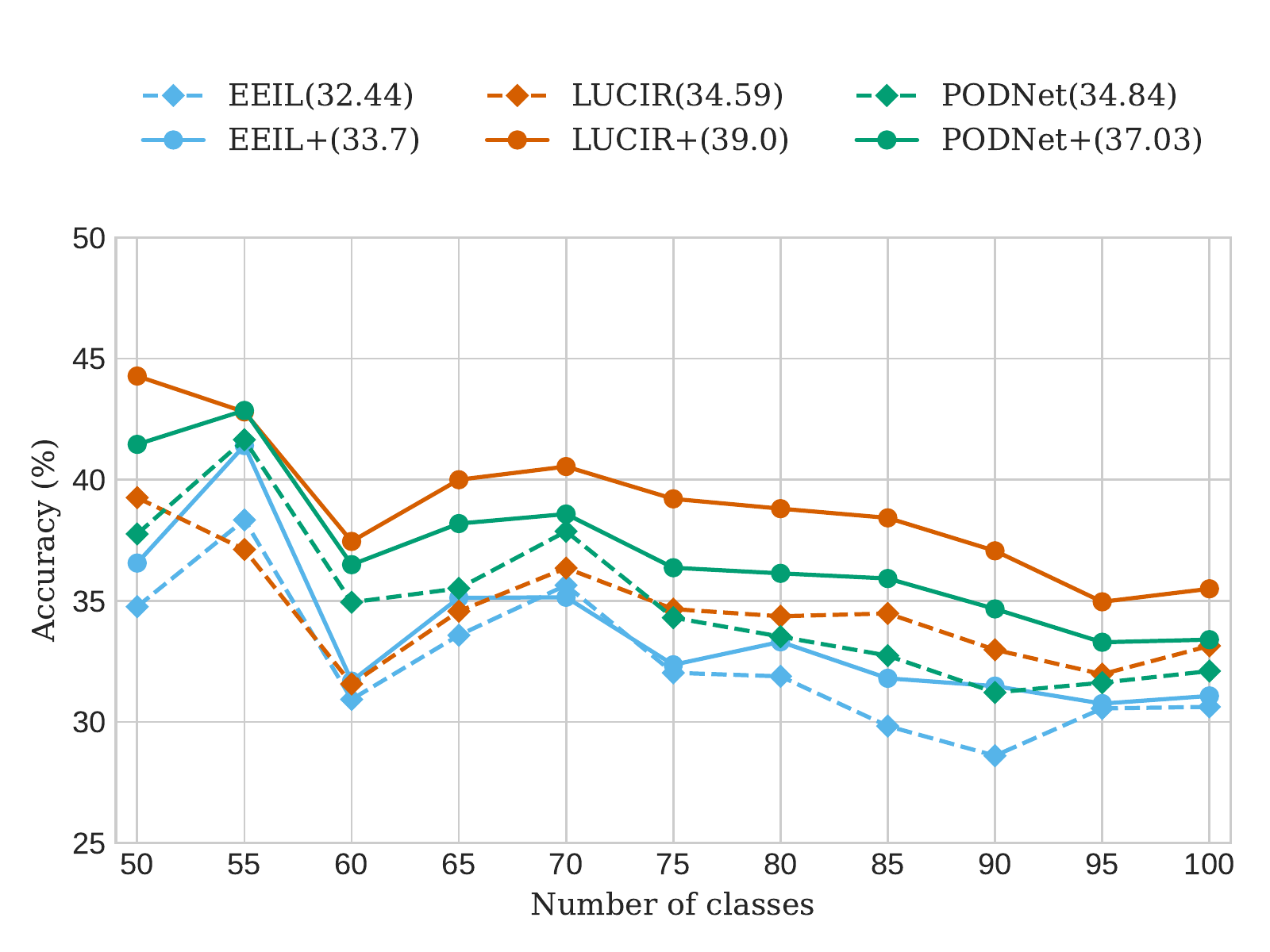}
\subcaption{$\rho$ = 0.01 Shuffled LT}
\end{minipage}
\begin{minipage}[b]{0.32\linewidth}
\centering
\includegraphics[width=\textwidth]{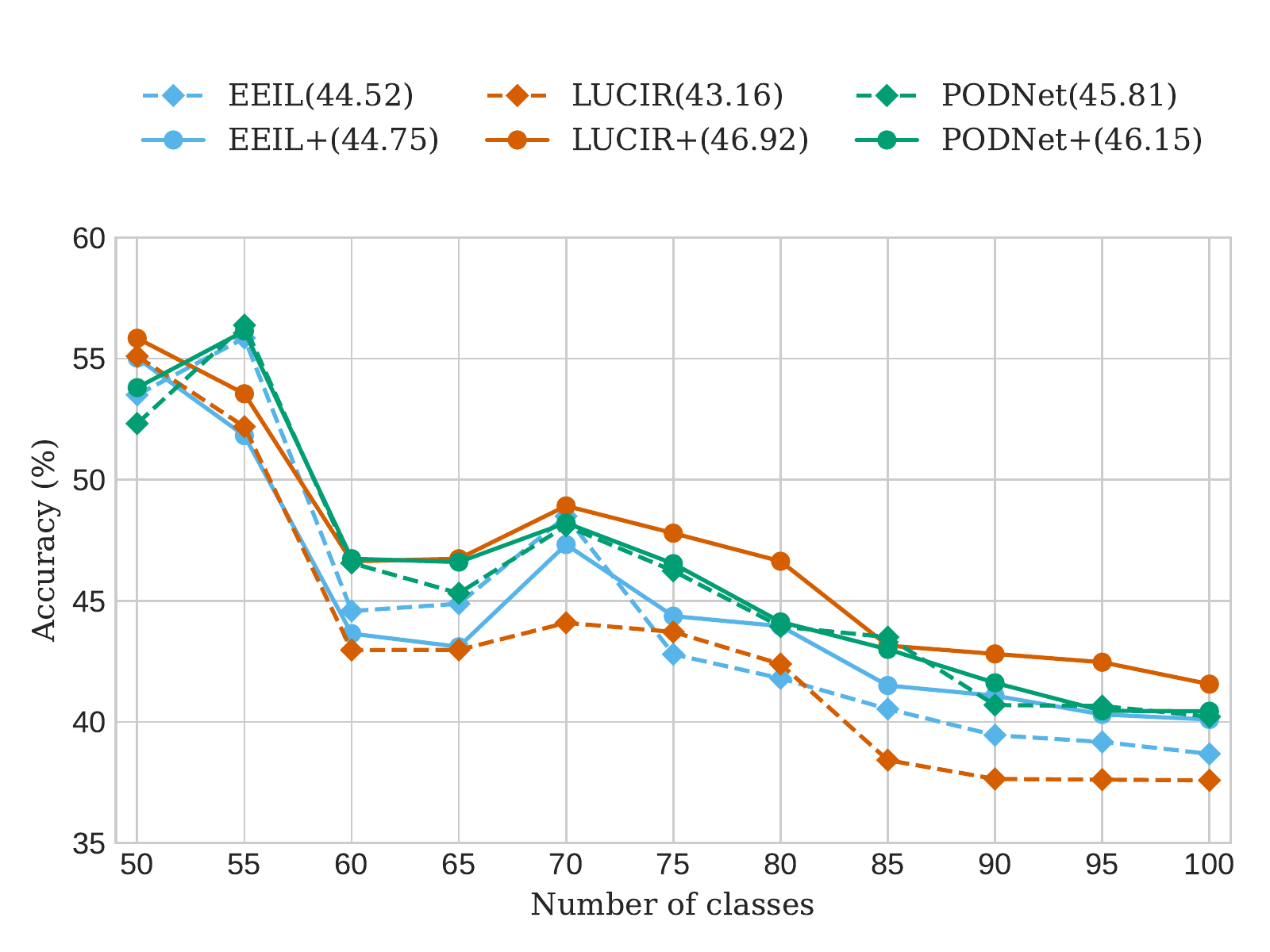}
\subcaption{$\rho$ = 0.05 Shuffled LT}
\end{minipage}
\begin{minipage}[b]{0.32\linewidth}
\centering
\includegraphics[width=\textwidth]{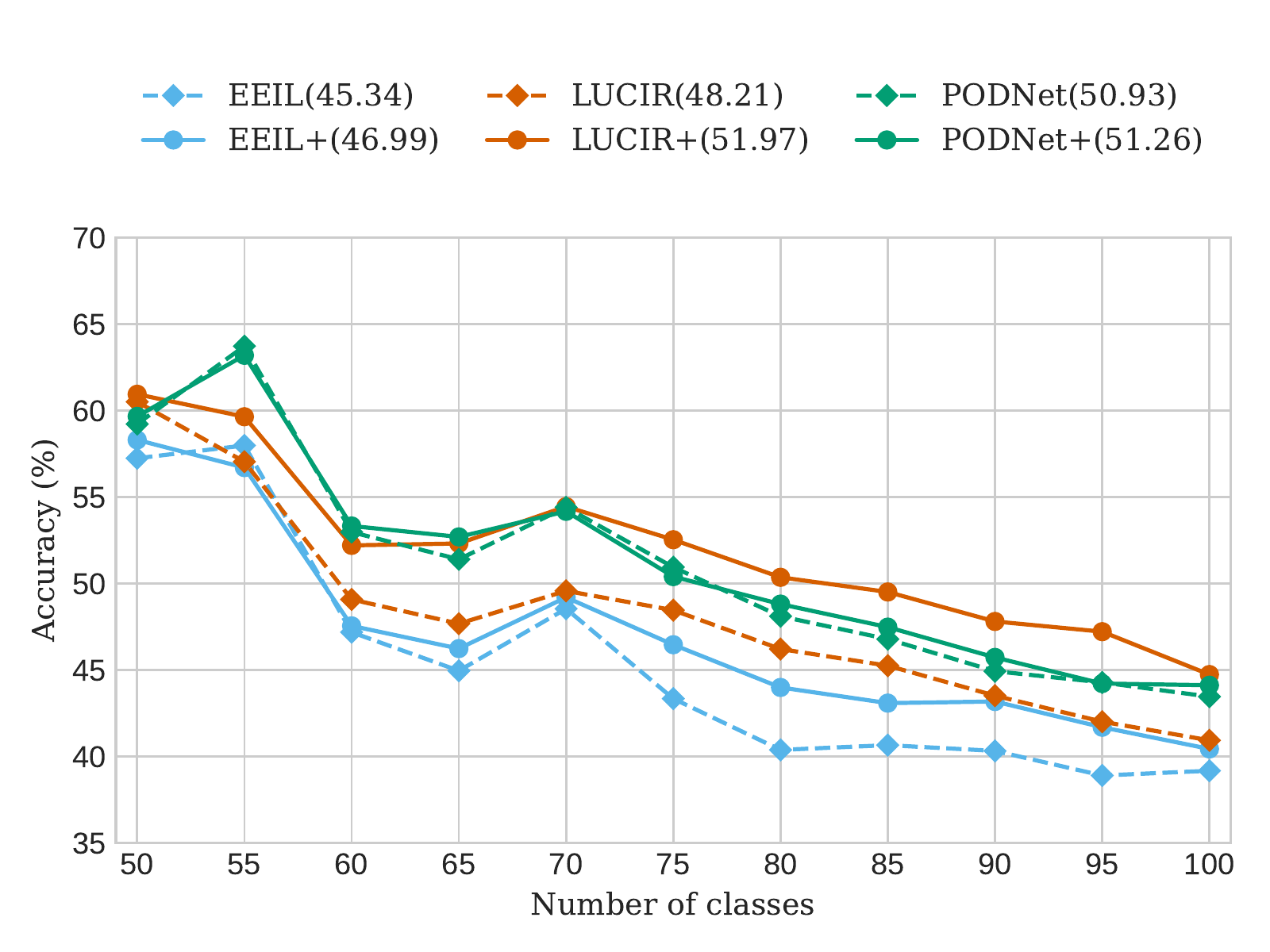}
\subcaption{$\rho$ = 0.1 Shuffled LT}
\end{minipage}
\caption{Average accuracy on CIFAR-100 dataset for different imbalance ratios. The top row is for Ordered LT-CIL and the bottom row for Shuffled LT-CIL. The $+$ suffix indicates our two-stage method applied to the corresponding baseline.
}
\label{fig:imb_ratio}
\end{figure*}

\subsection{Ablation study}
\label{sec:ablation}

\begin{table}[t]
\setlength{\tabcolsep}{5pt}{
\begin{tabular*}{230pt}{cc|ccc}
\toprule
Fix $h_{1:t-1}$ & LWS & \multicolumn{1}{l}{Conventional} & \multicolumn{1}{l}{Ordered} & \multicolumn{1}{l}{Shuffled} \\
\midrule
         &     & 26.28                                & 33.54                              & 23.41                               \\
\checkmark        &     & 60.00                                   & 43.52                              & 37.38                               \\
         & \checkmark   & 60.28                                & 44.45                              & 38.13                               \\
\checkmark       & \checkmark   & \textbf{60.57}                                & \textbf{45.73}                              & \textbf{39.01}      \\
\bottomrule
\end{tabular*}
}
\caption{Ablation study on effectiveness of different components. $h_{1:t-1}$ denotes the classification heads up to task $t-1$.}
\label{tab:lws}
\end{table}

\begin{figure*}[t]
\begin{minipage}[b]{0.32\linewidth}
\centering
\includegraphics[width=\textwidth]{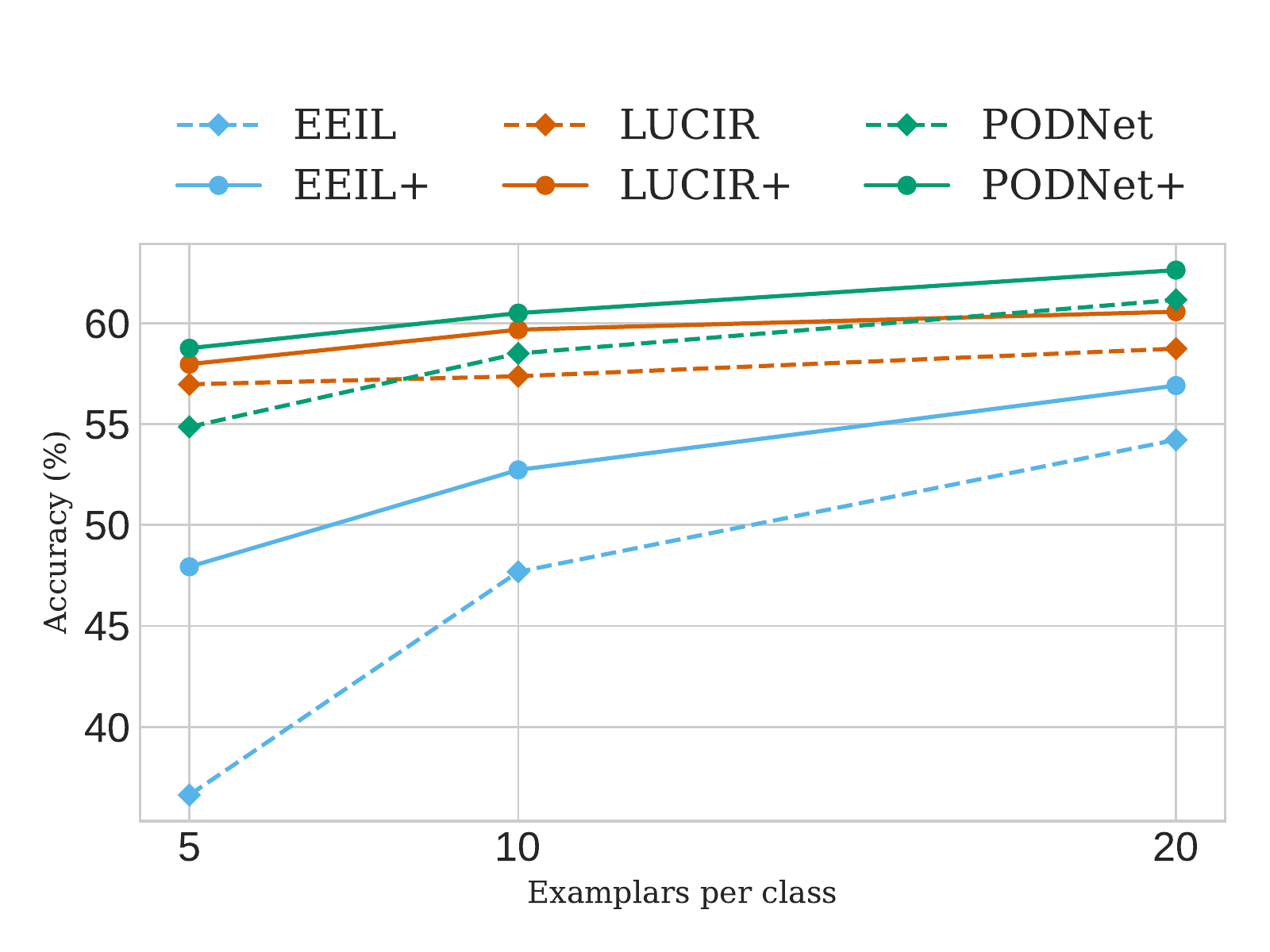}
\subcaption{Conventional CIL}
\end{minipage}
\begin{minipage}[b]{0.32\linewidth}
\includegraphics[width=\textwidth]{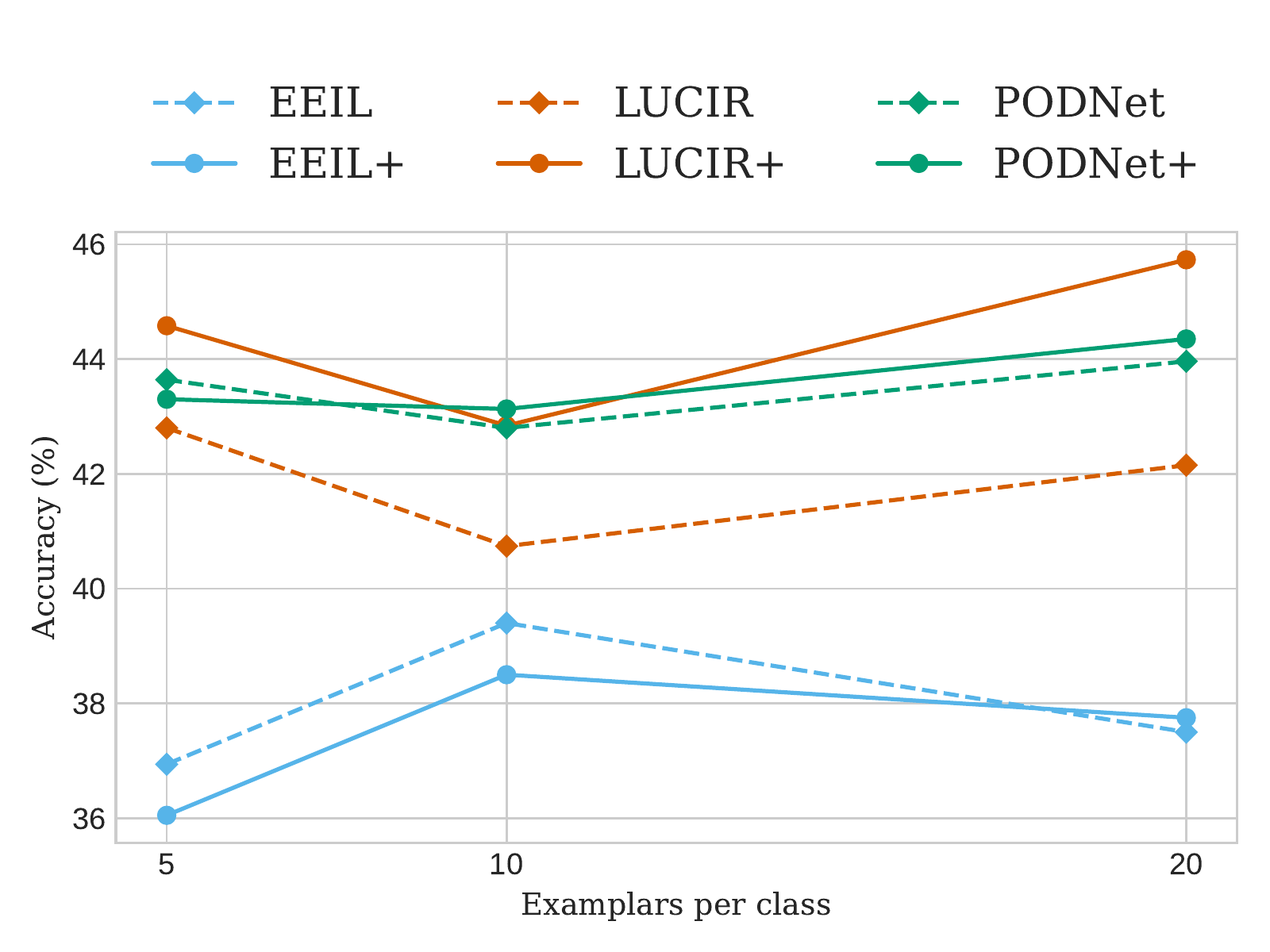}
\subcaption{Ordered LT-CIL}
\end{minipage}
\begin{minipage}[b]{0.32\linewidth}
\centering
\includegraphics[width=\textwidth]{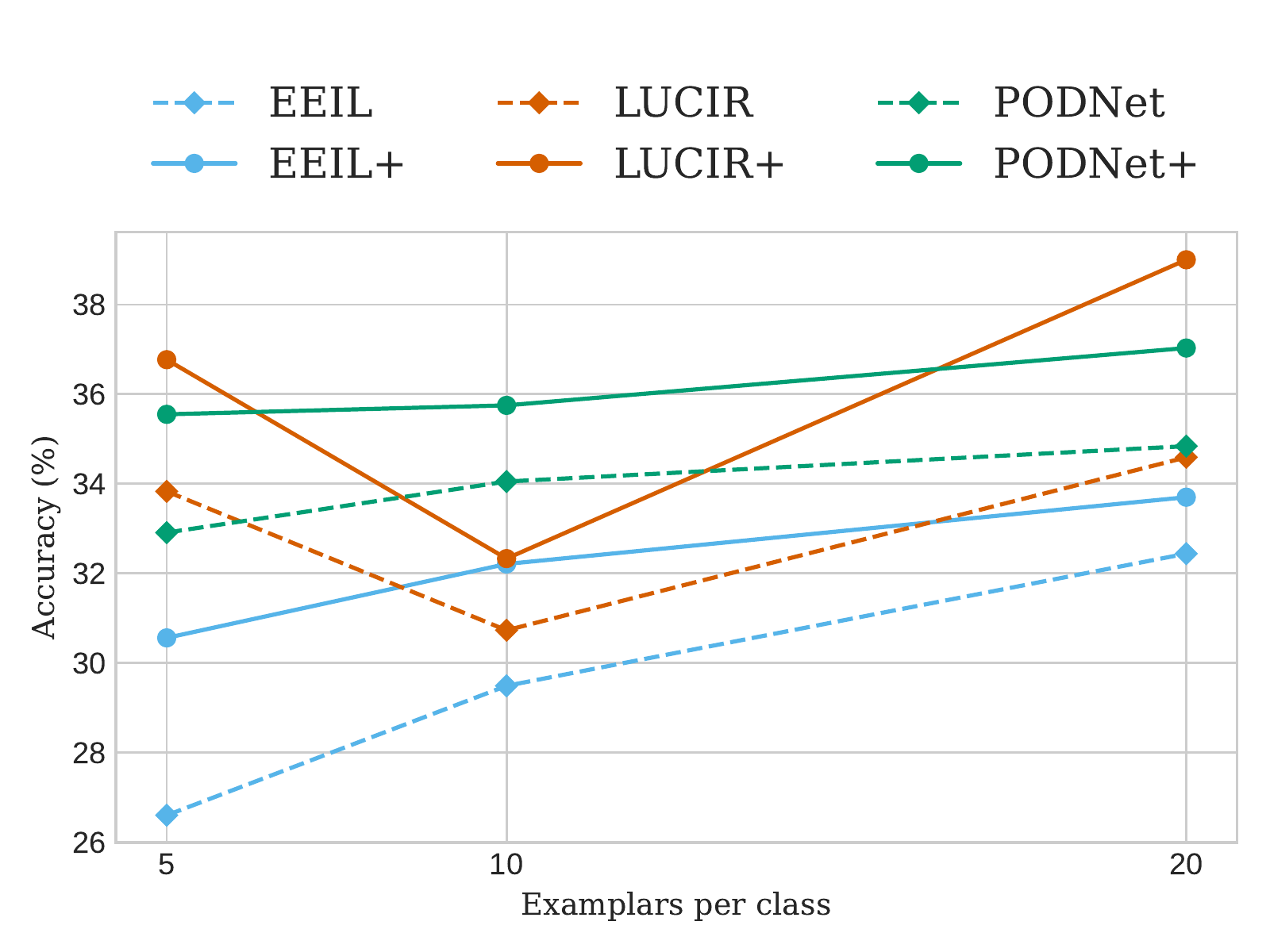}
\subcaption{Shuffled LT-CIL}
\end{minipage}
\caption{Average incremental accuracy on CIFAR-100 for different scenarios as a function of stored exemplars. The $+$ suffix indicates our two-stage method applied to the corresponding baseline.
}
\label{fig:ablation_memory}
\end{figure*}

\paragraph{Ablation on imbalance ratio.} In this section we analyze three different imbalance ratios: $\rho=0.01$, $\rho=0.05$ and $\rho=0.1$. The smaller the ratio, the more skewed the distribution. In Fig.~\ref{fig:imb_ratio} we give results for three different baselines (EEIL, LUCIR and PODNET) and our two-stage approach applied to them (EEIL+, LUCIR+ and PODNET+). As seen in Fig.~\ref{fig:imb_ratio}(a-c), for the Ordered LT-CIL scenario PODNET surpasses LUCIR by a larger margin as imbalance ratio $\rho$ increases. However, LUCIR obtains the best performance when $\rho=0.01$ but is worse than PODNET by a large margin when $\rho=0.1$. Overall, our two-stage method consistently boosts accuracy of most methods, especially for LUCIR+. For the Shuffled LT-CIL scenario in Fig.~\ref{fig:imb_ratio} (d-f), we see that PODNET outperforms LUCIR for all three $\rho$. The proposed two-stage method further improves performance, especially for LUCIR, resulting in LUCIR+ with the best overall performance. More results on ImageNet-Subset can be found in the supplementary material.

\paragraph{Ablation on exemplar memory size.} We evaluate different methods with 5, 10, and 20 exemplars per class. As expected, we see in Fig.~\ref{fig:ablation_memory}(a) that in the conventional CIL setting increasing exemplars results in better performance. However, for Ordered LT-CIL in Fig.~\ref{fig:ablation_memory}(b) we see that LUCIR and our two-stage method LUCIR+ both drop when increasing from 5 to 10 exemplars, but recover with 20 exemplars. For both EEIL and EEIL+, the best performance is obtained with 10 exemplars, which may be due to the long-tailed distribution of the final tasks. Both PODNET and PODNET+ obtains better performance with more exemplars. Similarly, for Shuffled LT-CIL in Fig.~\ref{fig:ablation_memory}(c) performance of both EEIL and PODNET increases with more exemplars, but LUCIR drops at 10 exemplars.

\paragraph{Effectiveness of different components.} We ablate the two main components of fixing previous head $h_{1:t-1}$ until $t-1$ task and using LWS in the second stage. As seen from Table~\ref{tab:lws}, without using either component the performance is very poor in all three scenarios. Both fixing the previous head $h_{1:t-1}$ and using LWS significantly boost accuracy, in particular for the conventional scenario which is up to 2.5 times higher. Using both components results in the best overall performance.

\begin{figure}[t]
    \begin{minipage}[b]{0.35\linewidth}
    \centering
    \includegraphics[width=\textwidth]{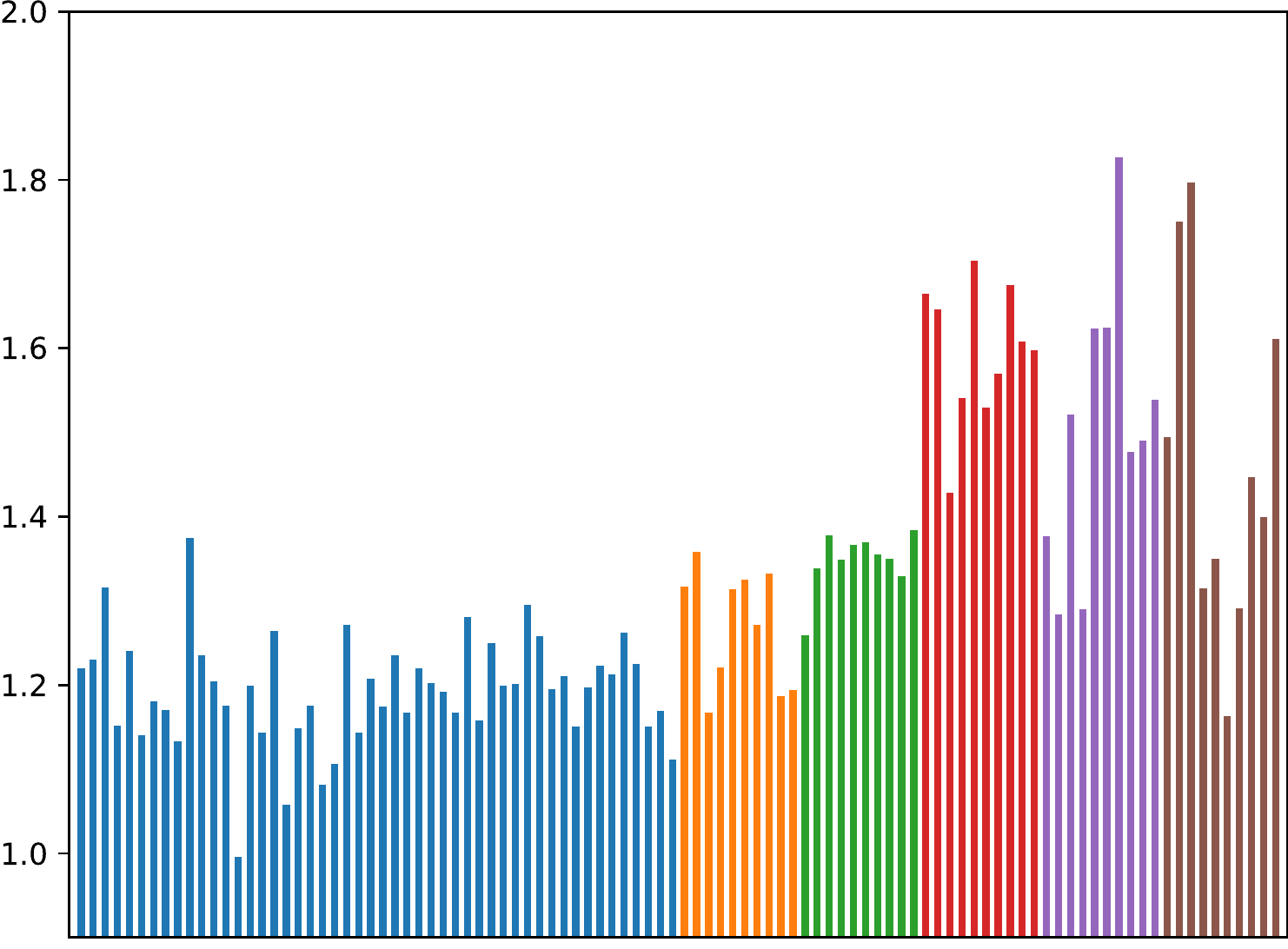}
    \subcaption{Ordered Long-tailed CIL}
    \end{minipage}
    \begin{minipage}[b]{0.35\linewidth}
    \centering
    \includegraphics[width=\textwidth]{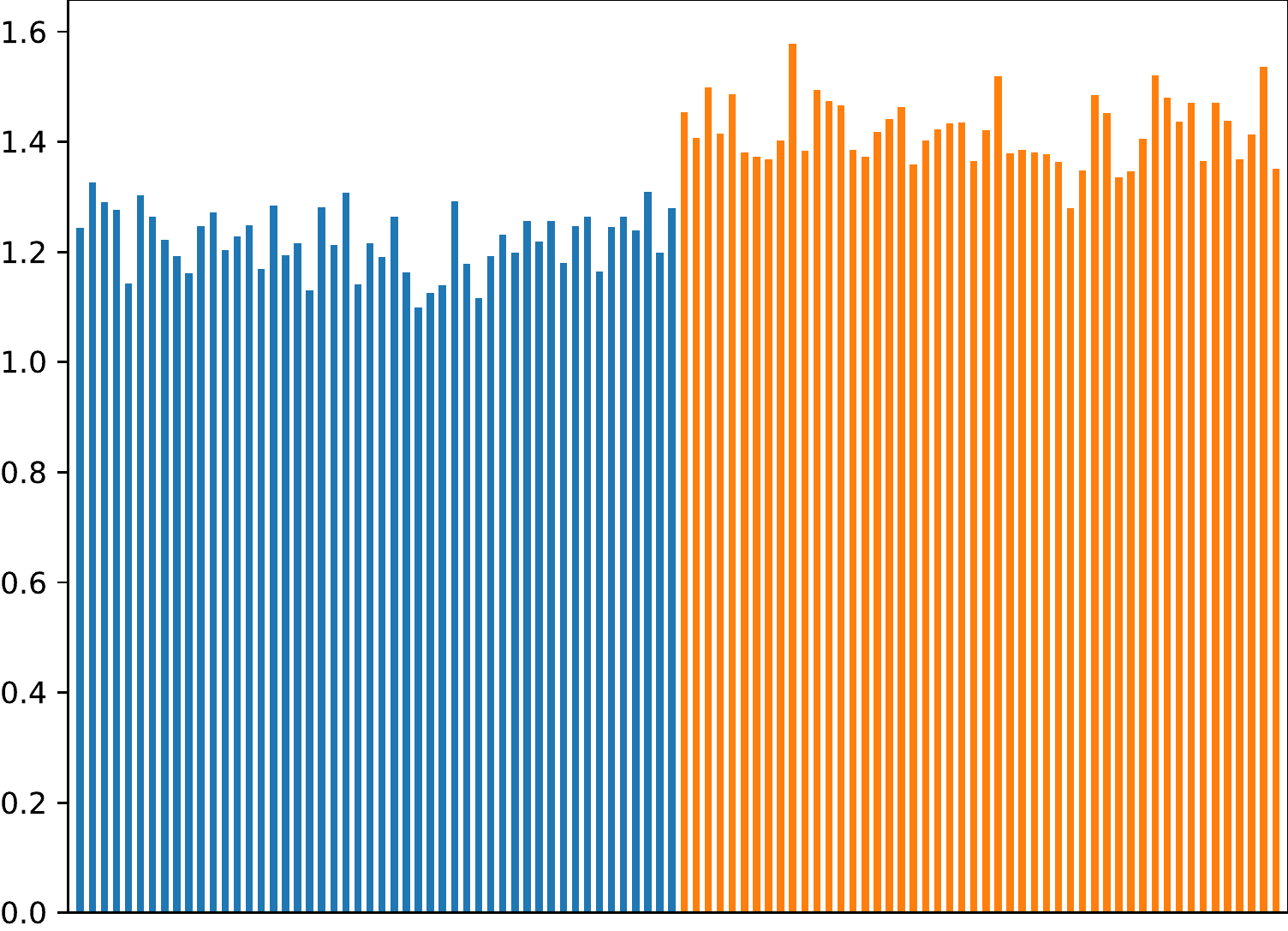}
    \subcaption{Conventional CIL}
    \end{minipage}
        \caption{LWS weights for Ordered LT-CIL (6 tasks) and conventional CIL scenarios (2 tasks) on ImageNet-Subset. Different colors indicate the weights for different tasks.}
    \label{fig:lws}
\end{figure}

\begin{table}[tb]\Large
\setlength{\tabcolsep}{1pt}{
\resizebox{0.8\columnwidth}{!}{%
    \begin{tabular*}{395pt}{l|cc|cc|cc}
    \toprule
                 & EEIL  & EEIL+ & LUCIR & LUCIR+         & PODNet & PODNet+        \\ \midrule
    Conventional & 51.07 & 53.24 & 52.02 & 56.25          & 58.42  & \textbf{60.47} \\
    Ordered      & 39.47 & 40.03 & 38.31 & \textbf{43.65} & 41.47  & 40.78          \\
    Shuffled     & 32.20  & 34.68 & 30.68 & \textbf{37.38} & 35.75  & 37.09          \\ \bottomrule
    \end{tabular*}}
}
\caption{Average accuracy on long sequences of 25 tasks for three different scenarios on CIFAR-100. The $+$ suffix indicates our two-stage method applied to the corresponding baseline.}
\label{tab:long}
\end{table}

\paragraph{Long task sequences.} In this experiment we evaluate on a longer sequence of 25 tasks in CIFAR-100 for all three scenarios. As we see in Table~\ref{tab:long}, our method improves over all baselines in this more challenging setting.LUCIR gains the most from our two-stage approach, with performance increasing by up to 6.7. More results on ImageNet-Subset can be found in the supplementary material.

 \paragraph{Analysis of LWS layer.} In Fig.~\ref{fig:lws} we plot the weights of the LWS layer for LUCIR+ after learning the last task. Since the classes in the end of training consist of fewer and fewer samples, the LWS is capable of learning larger weights for them to balance with previous classes. In the conventional CIL scenario, the LWS weight for the current task is significantly larger than the older one, which can help predict correct labels for current classifier without modifying the feature representations too much in a two-stage framework, and thus reducing the forgetting of previous knowledge.

\begin{figure}[t]
\begin{minipage}[b]{0.32\linewidth}
\centering
\includegraphics[width=\textwidth]{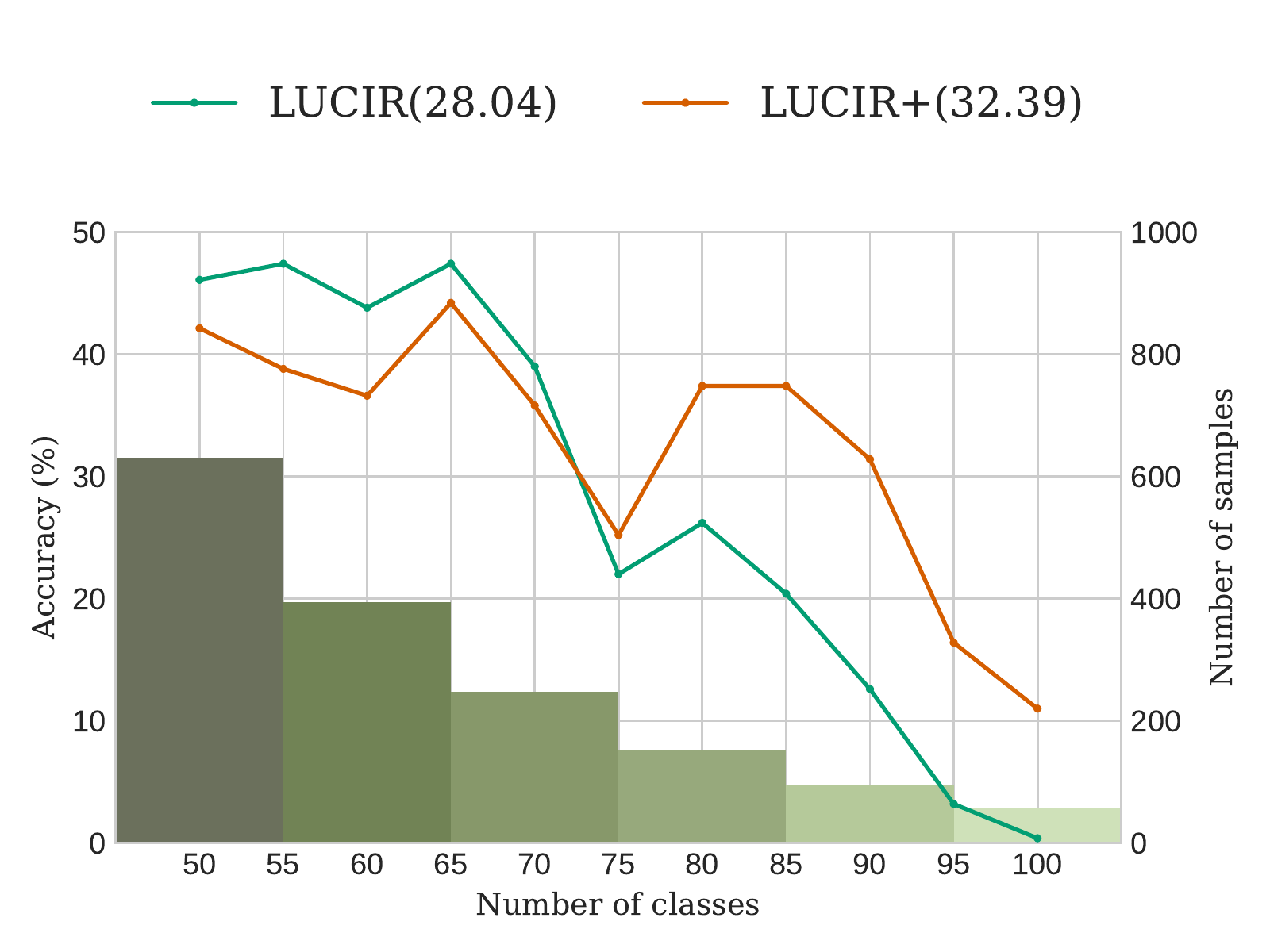}
\subcaption{LUCIR}
\end{minipage}
\begin{minipage}[b]{0.32\linewidth}
\centering
\includegraphics[width=\textwidth]{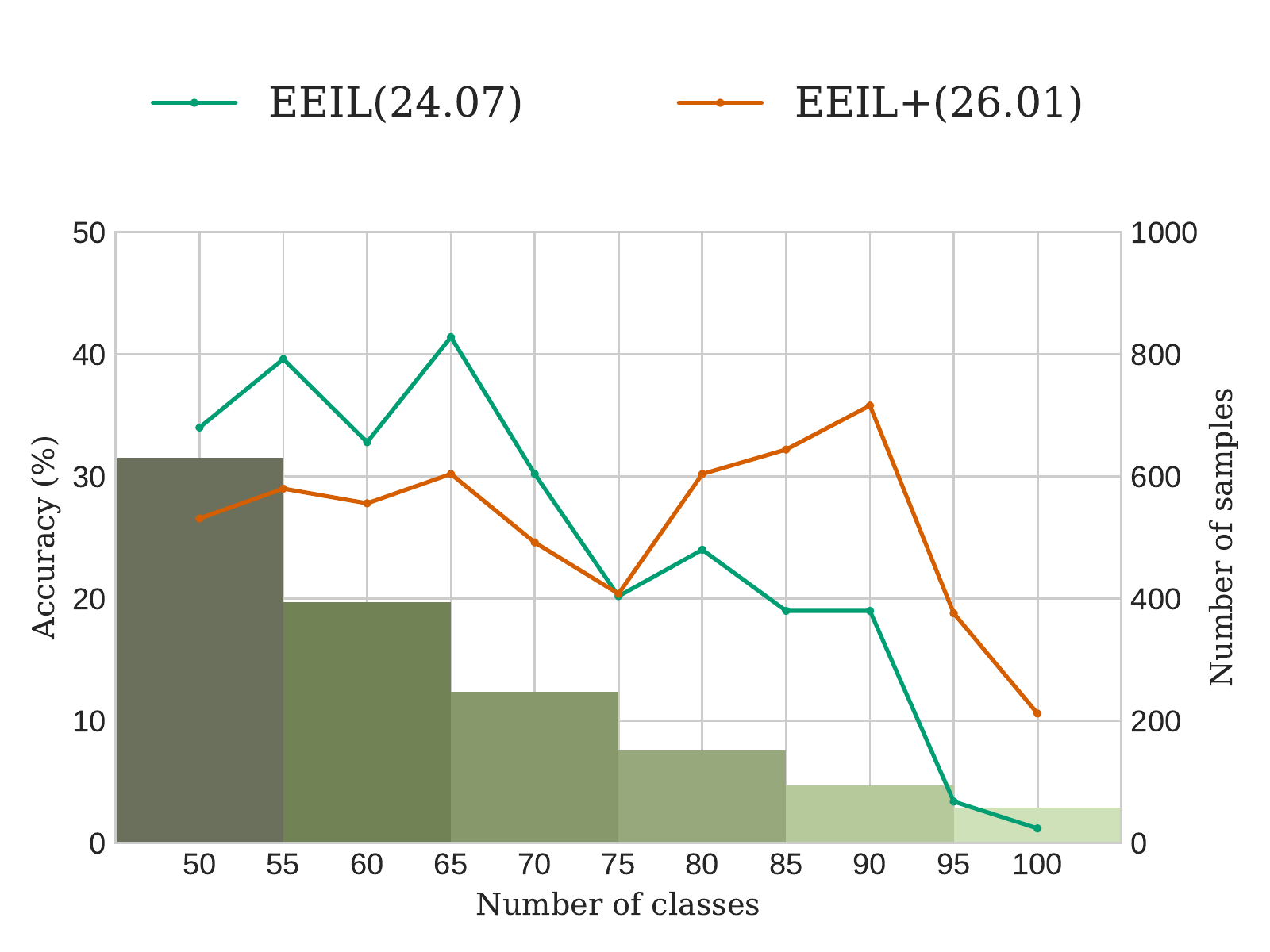}
\subcaption{EEIL}
\end{minipage}
\begin{minipage}[b]{0.32\linewidth}
\centering
\includegraphics[width=\textwidth]{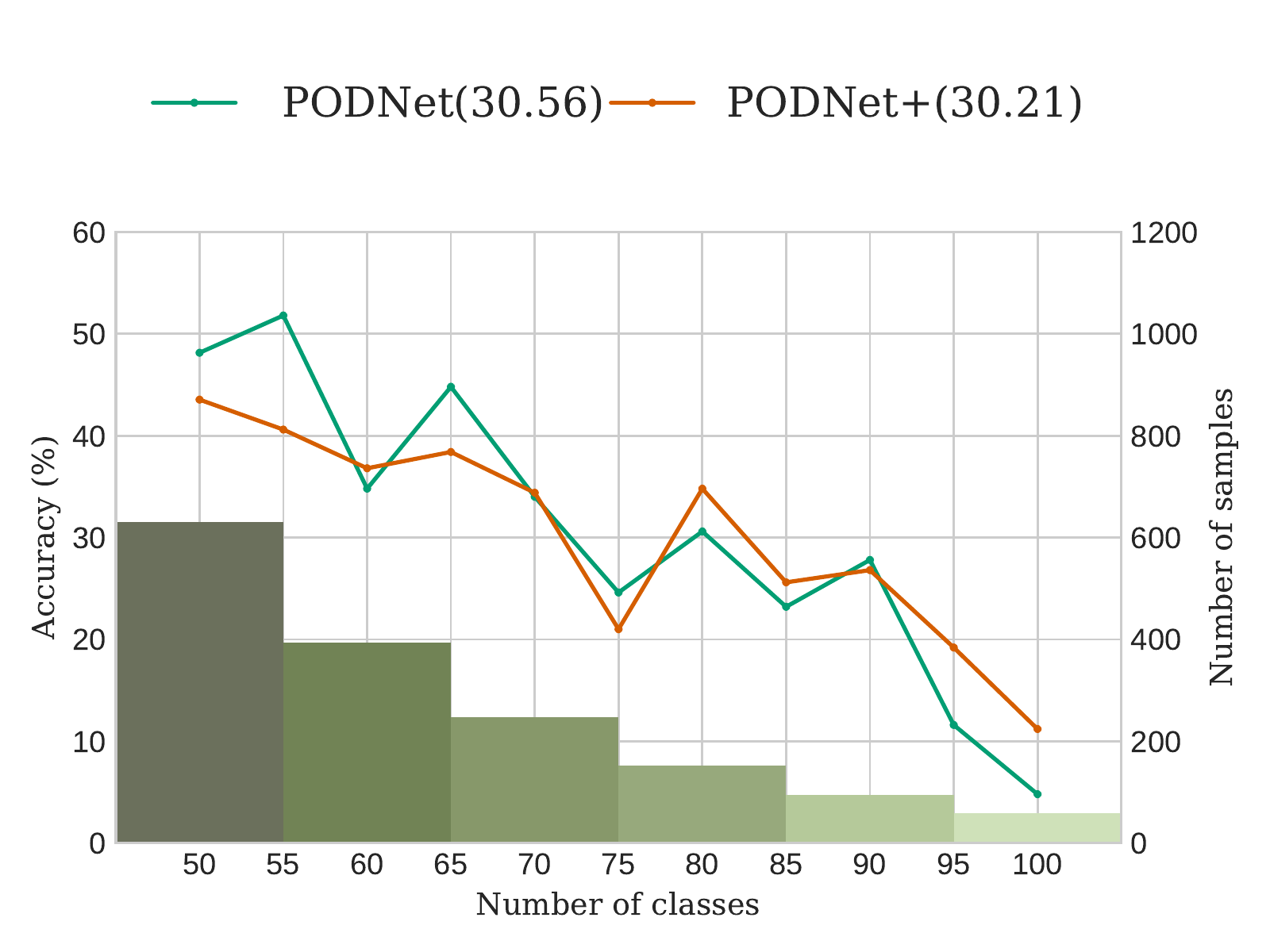}
\subcaption{PODNET}
\end{minipage}
\caption{The average accuracy curves with data distribution bars at the last task for Ordered LT-CIL on CIFAR100.
}
\label{fig:bar}
\end{figure}

\paragraph{Visualization of effectiveness of our method.} 
As seen in Fig.~\ref{fig:bar}, different methods and their corresponding two-stage versions are evaluated after learning the last task (5-tasks). It is clear that LUCIR+ and EEIL+ can significantly boost the performance of tail classes by losing relatively less for the head classes. PODNET+ improves over PODNET by a small margin for the tail classes with less drops for head classes.

\begin{figure}[t]
\begin{minipage}[b]{0.45\linewidth}
\centering
\includegraphics[width=\textwidth]{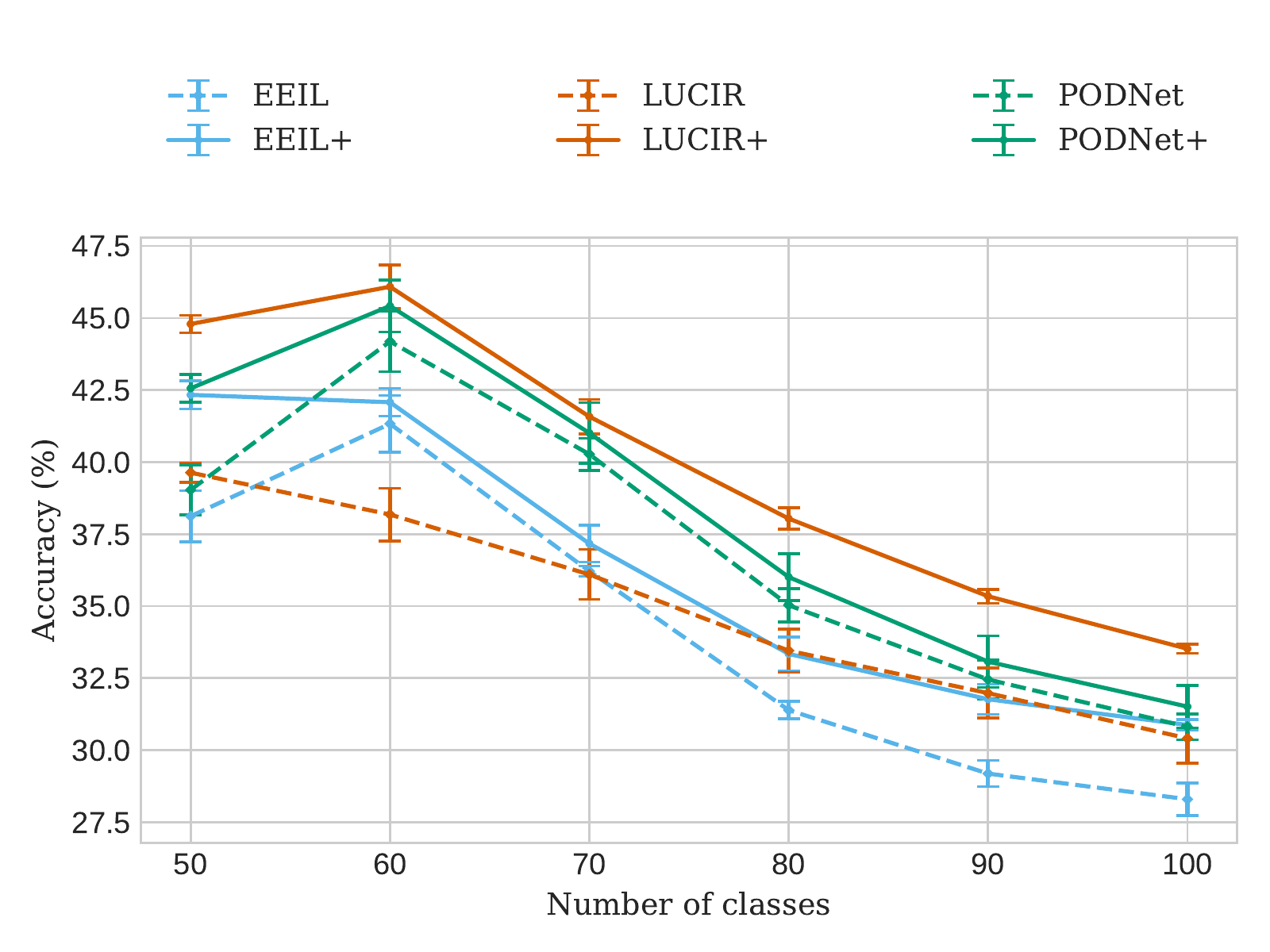}
\subcaption{5-task setting}
\end{minipage}
\begin{minipage}[b]{0.45\linewidth}
\centering
\includegraphics[width=\textwidth]{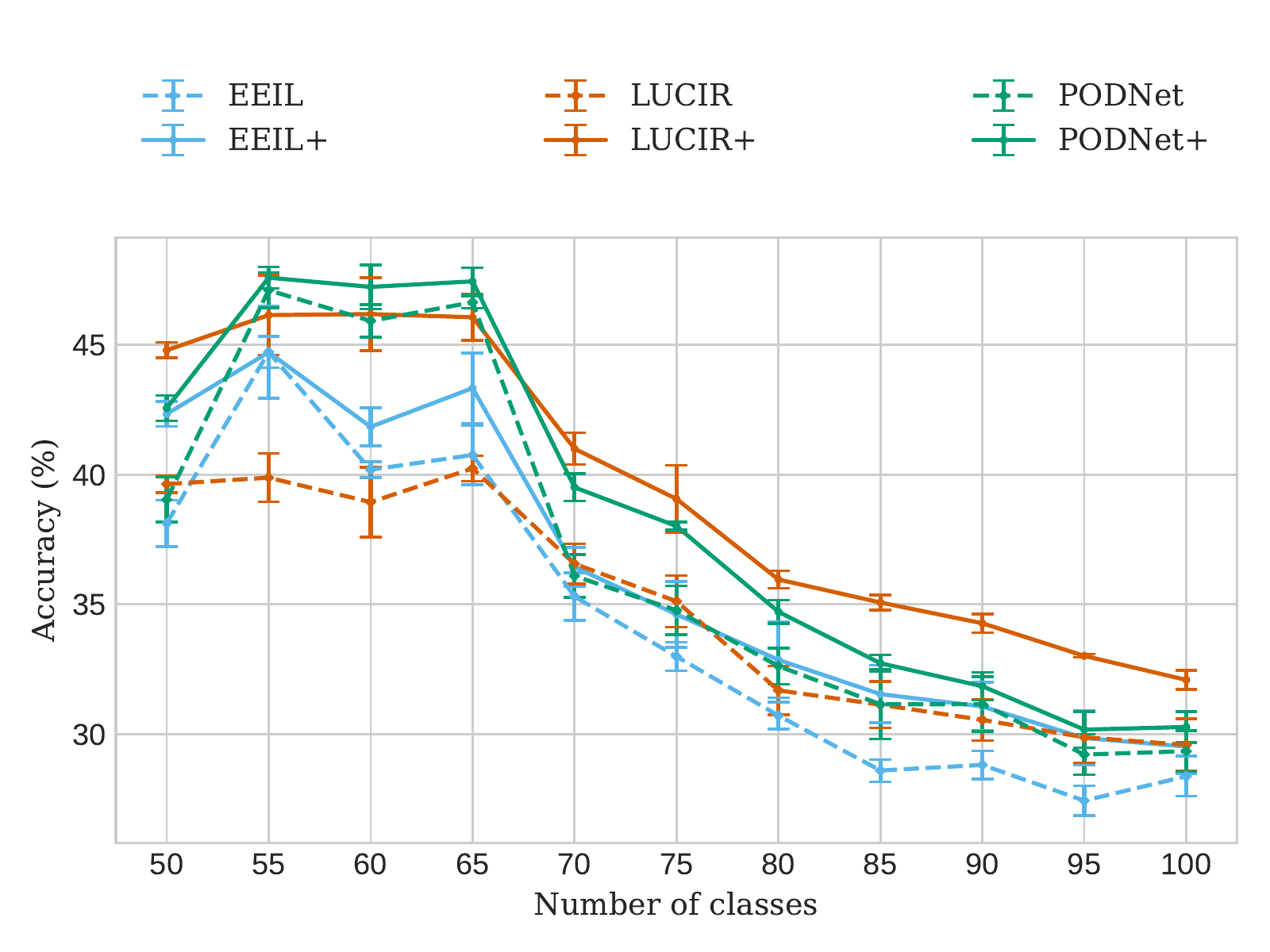}
\subcaption{10-task setting}
\end{minipage}
\caption{Average accuracy for Shuffled LT-CIL with multiple random seeds on CIFAR-100. Error bars are shown at each task.
}
\label{fig:ablation_seed}
\end{figure}

\paragraph{Ablation on class order.} To verify the robustness of our approach, we ran experiments with four different random seeds in the Shuffled LT-CIL scenario on CIFAR-100.As shown in Fig.~\ref{fig:ablation_seed}, the compared methods behave differently but the overall trend for all methods are clear. LUCIR remains on the top in the last several tasks. All methods with the proposed two-stage strategy improve over the baselines in both settings. 

%% file: conclusion.tex
In this paper we proposed two novel scenarios for class incremental learning over long-tailed distributions (LT-CIL). Ordered LT-CIL considers the case where subsequent tasks contain consistently fewer samples than previous ones. Shuffled LT-CIL, on the other hand, refers to the case in which the degree of imbalance for each task is different and randomly distributed.
Our experiments demonstrate that the existing state-of-the-art in CIL is significantly less robust when applied to long-tailed class distribution. To address the problem of LT-CIL, we propose a two-stage method with a learnable weight scaling layer that compensates for class imbalance. Our approach significantly outperforms the state-of-the-art on CIFAR-100 and ImageNet100 with long-tailed class imbalance. Our two-stage approach is complimentary to existing methods for CIL and can be easily and profitably integrated into them.
We believe that our work can serve as a test bed for future development of long-tailed class incremental learning.